\pdfoutput=1

\documentclass[11pt]{article}

\usepackage[]{acl}

\usepackage{times}
\usepackage{latexsym}

\usepackage[T1]{fontenc}

\usepackage[utf8]{inputenc}

\usepackage{microtype}

\usepackage{subcaption}
\usepackage{graphicx}
\usepackage{dirtytalk}
\usepackage{booktabs}
\usepackage{subcaption}
\usepackage{multirow}
\usepackage{arydshln}
\usepackage{comment}
\usepackage{longtable}
\usepackage{supertabular}

\usepackage{cleveref}
\crefformat{section}{\S#2#1#3}
\crefformat{subsection}{\S#2#1#3}
\crefformat{subsubsection}{\S#2#1#3}

\usepackage{xcolor}
\definecolor{lightergreen}{RGB}{26, 153, 0} 
\definecolor{lighteryellow}{RGB}{255, 198, 26}
\definecolor{lighterred}{RGB}{204, 0, 0}

\title{Knowledge of Knowledge: \\
\textit{Exploring Known-Unknowns Uncertainty with Large Language Models}}

\author{
Alfonso Amayuelas$^{1}$,  Kyle Wong$^1$, Liangming Pang$^1$, Wenhu Chen$^{2,3}$, William Wang$^1$ \\
$^1$University of California, Santa Barbara \\
$^2$University of Waterloo \\
$^3$Vector Institute \\
\texttt{\{amayuelas, knw, liangmingpan\}@ucsb.edu} \\
\texttt{wenhuchen@waterloo.ca, william@cs.ucsb.edu}
}

\begin{document}
\maketitle
\begin{abstract}
This paper investigates the capabilities of Large Language Models (LLMs) in the context of understanding their knowledge and uncertainty over questions. Specifically, we focus on addressing \textit{known-unknown} questions, characterized by high uncertainty due to the absence of definitive answers. To facilitate our study, we collect a new dataset with \textbf{K}nown-\textbf{U}nknown \textbf{Q}uestions (KUQ) and establish a categorization framework to clarify the origins of uncertainty in such queries. Subsequently, we examine the performance of open-source LLMs, fine-tuned using this dataset, in distinguishing between known and unknown queries within open-ended question-answering scenarios. The fine-tuned models demonstrated a significant improvement, achieving a considerable increase in F1-score relative to their pre-fine-tuning state. Through a comprehensive analysis, we reveal insights into the models' improved uncertainty articulation and their consequent efficacy in multi-agent debates. These findings help us understand how LLMs can be trained to identify and express uncertainty, improving our knowledge of how they understand and express complex or unclear information.
\end{abstract}

\section{Introduction}
\let\thefootnote\relax\footnotetext{\begin{flushleft}Source code and data available on: \href{https://github.com/amayuelas/knowledge-of-knowledge}{https://github.com/amayuelas/knowledge-of-knowledge}\end{flushleft}
}

\vspace{5pt}
\begin{center}
\begin{tabular}{p{0.7\linewidth}}

\say{\textit{To know what you know and to know what you do not know, that is true knowledge}  

\vspace{4pt}
\raggedleft
--- The Analects of Confucius}
\vspace{4pt}
\end{tabular}
\end{center}

\begin{table}[]
\scalebox{0.85}{
\begin{tabular}{c|c}
\textbf{Known Knowns} & \textbf{Known Unknowns} \\
\begin{tabular}[c]{@{}c@{}}Things we are aware of \\ and understand\end{tabular}  &  \begin{tabular}[c]{@{}c@{}}Things we are aware of \\ but do not understand\end{tabular} \\
\textit{\begin{tabular}[c]{@{}c@{}} e.g. What's the boiling \\ temperature of water?\end{tabular}} & 
\textit{\begin{tabular}[c]{@{}c@{}} e.g How many planets \\ are there in the universe?\end{tabular}} \\
\hline
\textbf{Unknown Knowns} & \textbf{Unknown Unknowns} \\ 
\begin{tabular}[c]{@{}c@{}}Things we understand\\ but are not aware of\end{tabular}  & \begin{tabular}[c]{@{}c@{}}Things we are neither\\ aware of nor understand\end{tabular} \\
\textit{\begin{tabular}[c]{@{}c@{}} e.g. How to tell the \\ stomach to digest?\end{tabular}} & \textit{\begin{tabular}[c]{@{}c@{}} e.g How does gravity work? \\ (before it was discovered) \end{tabular}}
\end{tabular}
}
\caption{Quadrant of Knowledge. Taxonomy of the different kinds of knowledge we can ask about, popularized by US Secretary of Defense Donald Rumsfeld. We focus on investigating Known-Unknowns, questions for which we do not have an answer.}
\label{Table:knowledge-quadrant}
\vspace{-12pt}
\end{table}

Large Language Models (LLMs) have grown in size and capabilities \cite{chainOfThought} \cite{evaluating_llm_code}. Consequently, different works raise the question of what the models learn and know \cite{what_lms_know} and how they can express uncertainty \cite{model_uncertainty_in_words} \cite{navigating_the_grey_area}. 

We look at cognitive psychology, where \textit{metacognition} \cite{metacognition} is defined as the awareness and thoughts of one's own thought process. Do LLMs know what they know? And more importantly, are they aware of what they do not know? This is an important question to calibrate the certainty of their statements or prevent such language models from confidently generating false answers, commonly known as hallucinations.

Given the division of knowledge in Table  \ref{Table:knowledge-quadrant}, we pay special attention to \textit{Known-Unknowns}. These are questions that do not have a definitive answer. The answers to such questions are often subjective and may even be unanswerable due to a lack of information or inherent complexity. As a result, the answers are considered to have high uncertainty levels. For example, \textit{If the Universe started at the Big Bang, what existed before then?}. Our goal is to understand how language models deal with these uncertain questions. 

We identify several reasons why questions may be unknown. In some cases, questions that ask about future events inherently lack definitive answers due to the uncertainty of future outcomes. In other cases, some questions address unresolved issues in science or history, for which conclusive answers are currently unavailable. Additionally, questions may contain incorrect assumptions, making them unanswerable as they are based on false premises. A comprehensive categorization of such questions is presented in Table \ref{Table:unknown_categories}.

The topic of \textit{known-unknown} questions has barely been studied in the area of Large Language Models. SelfAware \cite{selfaware} introduces the topic, but our work carries a detailed analysis of open-source models, how they can be fine-tuned, and a categorization of the questions with explanations. Below, we present the main research questions and contributions from our work:


\noindent $\bullet$ \textbf{Can open-source models differentiate between known and unknown questions?}. We introduce a dataset of \textbf{K}nown-\textbf{U}nknown \textbf{Q}uestions (\textbf{KUQ}) and evaluate it on the open-source LLama family of models. We show how the LLama models fall behind in this task when compared to GPT-3.5 and GPT-4, and introduce a fine-tuning strategy that brings them on par with these models.

\noindent $\bullet$ \textbf{How does fine-tuning improve the ability of open-source models to differentiate between known and unknowns?} Fine-tuning proves to be a good strategy for adding new abilities to open-source models. Identifying the question of uncertainty is one of them. However, we see a trade-off between this ability and correctly answering known questions.  We also show the generalization ability of these models on the self-aware dataset. 

\noindent $\bullet$ \textbf{Can a fine-tuned model on our KUQ dataset improve the results of a downstream task?} Understanding and expressing uncertainty has many potential applications. We show our fine-tuned model on KUQ can enhance the results of multiagent debate on some reasoning tasks.

\section{Related Work}

\paragraph{Language Models Knowledge.}
Since the beginning of the first pre-trained language models, some researchers have studied what information is stored in their weights and how we can extract that knowledge \cite{what_lms_know}, along with how confident the models are about their knowledge \cite{when_lms_know}. More recently, another work has explored whether LLMs can evaluate the validity of their claims \cite{llm_mostly_know}.  The question of what a model should know has also been explored in computer vision systems \cite{unknowns_computerVision}, with a human-in-the-loop process to investigate what the models really know and what they should know. These works look at the model knowledge, but we want to take a step further by providing questions that are uncertain by themselves. This question has initially been explored in SelfAware \cite{selfaware}.

\paragraph{Language Models Uncertainty.} \citep{uncertatintyNLP_review}
Modeling uncertainty has been a persistent challenge for the linguistics community. Uncertainty can be divided into \textit{epistemic uncertainty}, which refers to the model uncertainty, and \textit{aleatoric uncertainty}, which belongs to the data's inherent randomness. \citet{ambiguousQuestions} introduces the notion of \textit{denotional uncertainty} for the uncertainty contained in the meaning of the question. Denotional uncertainty is the area addressed in this work. Expressing uncertainty requires knowledge about one's own knowledge and the ability to define the level of confidence in one's response. Several studies have examined various approaches to expressing uncertainty \citep{semantic_uncertainty, cross_semantic_uncertainty, uncertainty_hedges}. In the area of Large Language Models, some works have quantified uncertainty \cite{uncertaintyQuantification}, and explored how they behave when expressing uncertainty \cite{navigating_the_grey_area} and how they can learn to express uncertainty in words \cite{model_uncertainty_in_words}. 

\section{Data}

In the evaluation domain, the significance of datasets containing known-unknown questions is paramount. Known-unknown questions are those that do not have definitive answers, such as \textit{"Are there other forms of intelligent life in the universe?"} or \textit{"Which year will the next financial crisis occur?"}. Upon surveying available resources, we identified a mere 46 known-unknown questions in the Big-Bench benchmark \cite{bigBench} and 1 evaluation dataset in SelfAware \cite{selfaware}. This limited quantity is insufficient for robust training + evaluation, prompting us to generate additional samples for a full-scale dataset. Our dataset is the first of its kind to include annotations on the questions pointing to the reason for uncertainty. 
The dataset is publicly available.

\begin{figure}
    \centering
    \includegraphics[width=\linewidth]{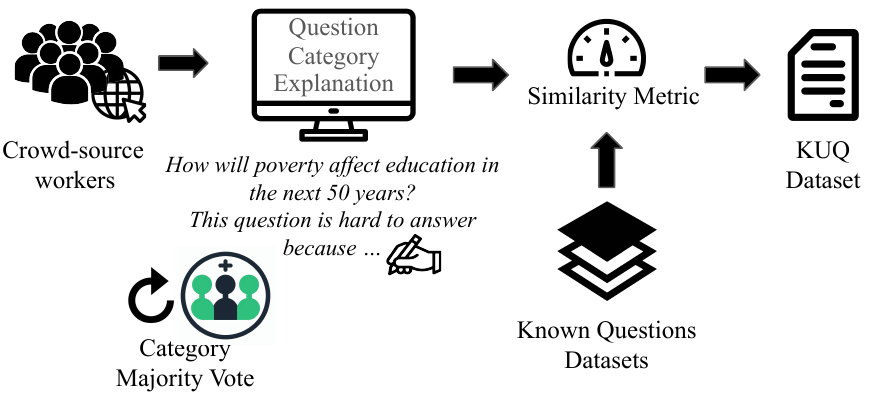}
    \caption{Data Collection Process: (1) Prompt crowd-source workers with Question Type, Category, and Explanation. (2) Confirm Category and Explanation with a Majority Vote of 3 workers. (3) Match Unknown Questions with Known Questions through Similarity Metric}
    \label{fig:dataset-generation}
\end{figure}

\begin{table}[]
\centering
\scalebox{0.85}{
\begin{tabular}{cc}
\toprule
\multicolumn{1}{c}{\textbf{Source}}       & \textbf{\#Questions} \\ \hline 
\multicolumn{2}{c}{\textbf{Unknown} (49.9\%)}                    \\ 
\multicolumn{1}{c}{Crowd-source} & 3437        \\ \hline
\multicolumn{2}{c}{\textbf{Known} (50.1\%)}                      \\ 
\multicolumn{1}{c}{Squad}        & 1928        \\ 
\multicolumn{1}{c}{TriviaQA}     & 854         \\ 
\multicolumn{1}{c}{HotPotQA}     & 665         \\ 
\multicolumn{1}{c}{\textit{Subtotal}}     & 3447         \\ \hline \hline
\multicolumn{1}{c}{\textbf{Total}}     & 6884 \\
\bottomrule
\end{tabular}
}
\caption{\textbf{K}nown-\textbf{U}nknown \textbf{Q}uestions (\textbf{KUQ}) Dataset Statistics: Number and source of questions.}
\label{Table:dataset_statistics}
\vspace{-12pt}
\end{table}

We have collected questions from crowd-source workers and generated a new dataset: \textbf{K}nown-\textbf{U}nknown \textbf{Q}uestions, \textbf{KUQ}. 
We have depicted the data collection process in Figure \ref{fig:dataset-generation} and explained question generation in \cref{Section:Data-Generation}.
Table \ref{Table:dataset_statistics} shows the number of questions generated per source. Hereafter, we refer to known-unknowns as unknown questions and known-knowns as known questions. 

In addition, we have identified several classes of unknown questions, shown in Table \ref{Table:unknown_categories}. These classes serve as a guideline on the source of uncertainty for each question. This dataset is more comprehensive than previous ones as it includes a larger set of categories and questions. For example, in SQuAD2 \cite{squad2}, they adversarially generate unanswerable questions. This differs from the general KUQ unknown questions, which are hard to answer because we cannot provide a correct answer, instead of the question being wrong by itself. We can argue similar cases for ControversialQA \cite{controversialQA}, which focuses on questions where there are controversial answers to the questions. Or AmbigQA \cite{ambigQA}, which focuses on questions that need clarification.  


\subsection{Data Generation}
\label{Section:Data-Generation}

\begin{table*}[]
\centering
\scalebox{0.83}{
\begin{tabular}{lllc}

\textbf{Categories}                       & \textbf{Explanation}                                                                                                                                      & \textbf{Example}          & \textbf{\#Questions}                                                       \\
\hline
\hline

\textbf{Future Unknown}                   &  
Questions about the future we cannot know  
& 
\begin{tabular}{@{}l@{}} 
\textit{What will be the top performing} \\ \textit{stock in 10 years?}
\end{tabular}
& 659 (19.2\%)
\\
\hline
\textbf{Unsolved Problem}
&  
\begin{tabular}{@{}l@{}} Questions about science, history or \\  problems that we don't know the answer to
\end{tabular}
&
\begin{tabular}{@{}l@{}} 
\textit{Is there a physics theory} \\ \textit{that can explain everything?}

\end{tabular}
& 437 (12.7\%)
\\
\hline
\textbf{Controversial} & 
\begin{tabular}{@{}l@{}} Subjective questions that have \\ different answers depending on the person
\end{tabular}

& \textit{How do you describe happiness?}
& 676 (19.7\%)
                                         \\

\hline

\textbf{w/ False Assumption}
   &  
\begin{tabular}{@{}l@{}} Questions that contain \\ a false assumption  or statement
\end{tabular}
& 
\begin{tabular}{@{}l@{}} \textit{Which city would hold the next} \\  \textit{Olympics if Detroit hadn't been elected?}
\end{tabular} 
& 520 (15.1\%)
\\

\hline

\textbf{Counterfactual}  & 
\begin{tabular}{@{}l@{}l@{}} Question based on a hypothetical \\scenario. It may ask about alternative \\ possibilities for past or future events.
\end{tabular}   &
\begin{tabular}{@{}l@{}} \textit{What would happen if the US} \\  \textit{had lost the Independence War?}
\end{tabular}
& 568 (16.5\%)
\\

\hline
\textbf{Ambiguous}    & 
\begin{tabular}{@{}l@{}l@{}} Questions that do not have an answer \\ because they are not specific enough \\ or they are incomplete
\end{tabular}
 & 
 \begin{tabular}{@{}l@{}}  \textit{What is the exact weight} \\  \textit{of a watermelon? }
\end{tabular}   
& 577 (16.8\%)
\\
\bottomrule
\end{tabular}
}
\caption{KUQ Unknown Questions Categories. It presents our categorization of unknown questions based on the source of the question uncertainty and the number of questions per category.}
\label{Table:unknown_categories}
\end{table*}

\textbf{Unknown Questions}. The unknown questions were carefully generated by crowd-sourced workers, an inherently difficult process. To ensure quality, we explained the concept of known-unknown questions to the workers and provided them with a category from Table \ref{Table:unknown_categories} along with examples. They were asked to generate a question and detail why its answer remains unknown. The tasks given to the workers are documented in Appendix \ref{App:Crowd-source-Dataset-Generation}.

\textbf{Known Questions}. The set of known questions has been selected to match the unknown questions. We have selected a set of well-known datasets with Question-Answer pairs: SQuAD \cite{squad}, TriviaQA \cite{triviaqa}, and HotPotQA \cite{hotpotqa}. From the pool of these 3 datasets together, known questions have been selected to match each of the unknown questions with SimCSE \cite{simcse}, a contrastive-learning framework used to find similar sentences.

\textbf{Categories}. Initially, we collected a diverse range of uncategorized questions from the web and via crowd-sourcing, which our team then analyzed and categorized based on identified common features. Some of these categories matched those in existing research, while others were new. In the final stage, we shared these categories with our crowd-sourced contributors, ensuring a clear understanding of the task and helping us collect a well-balanced set of questions across all categories.


\section{Methodology}

\subsection{Tasks}

We study 3 different tasks in this paper which are analyzed in the Experiments Section \cref{Section:Experiments}.

\noindent \textbf{1. Known vs Unknown}. The underlying idea of our work is the ability of open-source models to differentiate known and unknown questions in an open-ended question-answering scenario. Given the question, can the language model answer the question or express the question uncertainty otherwise? We also examine the ability of the models to differentiate the question categories introduced in Table \ref{Table:unknown_categories}.

\noindent \textbf{2. Effects of fine-tuning on KUQ}. We perform an analysis of the trade-offs of using fine-tuning to gain the skill to differentiate between known and unknown questions. 

\noindent \textbf{3. Downstream Application: Multiagent Debate}. The fine-tuned models on KUQ can be useful to improve downstream applications. In particular, we look into Multiagent Debate \cite{multiagent_debate}, where different versions of the language model discuss and compare their answers and thought processes over several rounds. Through this back-and-forth conversation, they work together to agree on a final answer on different knowledge and reasoning datasets. 

\subsection{Fine-tuning}

\begin{figure}[h]
    \centering
    \includegraphics[width=0.96\linewidth]{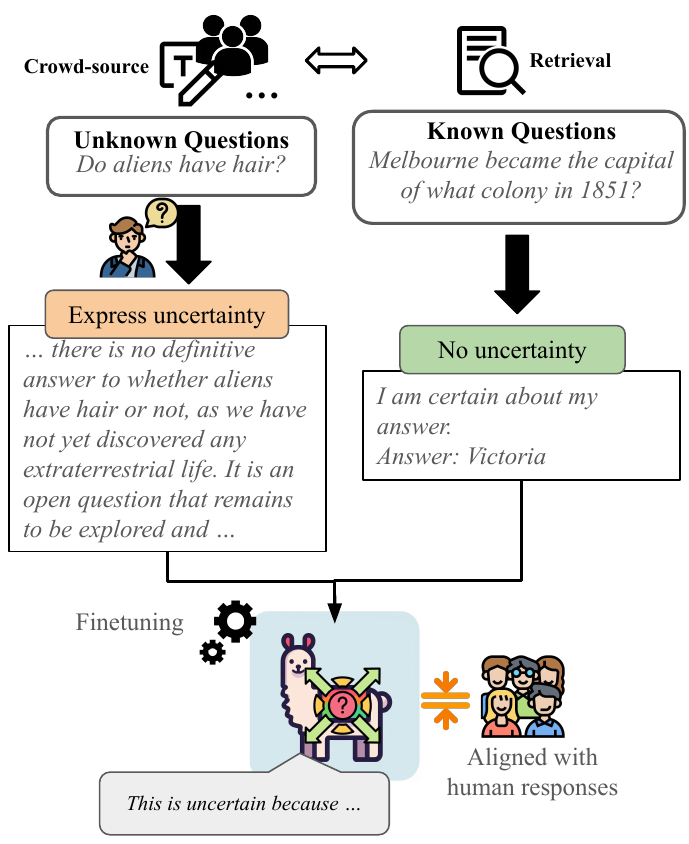}
    \caption{Fine-tuning Process with KUQ Dataset to elicit Question Uncertainty Understanding and Explanation. The fine-tuning process aligns model responses to human knowledge.}
    \label{fig:finetuing}
\end{figure}

In this work, we employed the KUQ dataset to fine-tune several open-source Large Language Models from the Llama-2 family. Our objective was to enhance their capabilities in expressing uncertainty when confronted with questions of an unknown or uncertain nature. The process is described in Figure \ref{fig:finetuing}

The dataset was specifically tailored for the fine-tuning process, incorporating either direct answers for known questions or expressions of uncertainty (see \cref{Section:KnownVsUnknown}) and category explanations (\cref{Appendix:Effects-Categories}) as provided by crowd-sourced workers. The format used was: \textit{\#\#\#Question:... \#\#\#Answer: \{Answer\} / The question may be unknown because...}.

We adopted LoRA \cite{lora} within the Peft framework from Huggingface, as this is a less resource-intensive fine-tuning approach. We conducted fine-tuning on the Llama-2 7B and 13B models, as well as their respective RLHF Chat versions \cite{llama2}, utilizing the Nvidia Titan RTX (24GB) and the RTX A6000 (48GB) graphics cards.

\subsection{Evaluation}
\label{Section:Evaluation}
The experiments conducted are centered around the open-ended Question-Answering scenario. In this setup, models are presented with questions and are expected to generate their answers. Our default approach is direct question-answering, unless specified otherwise. A key aim of our evaluation is to discern whether the text generated by the language models expresses uncertainty when responding to unknown questions. Additionally, we assess if these models can accurately provide the correct category after undergoing fine-tuning. This approach is based on methodologies established in previous work from SelfAware \cite{selfaware}.

We define a similarity function, $f_{sim}$ as a binary metric between the generated text ($t_i$) and some reference text ($ref_i$) to be 1 if they express the same content, or 0 if they do not. If the reference text is contained in the generated text or the similarity measured with SimCSE \cite{simcse} is higher than a threshold, $\tau$, the function returns 1.  

\begin{equation}
Sim_i = f_{sim}(text_i, ref_i) = \{0,1\}
\end{equation}

For differentiating between known and unknown questions, the reference texts are a predefined set of phrases that encompass general uncertainty. The full list can be found in 
Appendix \ref{App:List of Uncertainty Expressions}. For example, in the following generations, we would expect: 
\textit{It is difficult to predict the future, the reason why …} -> Contains Uncertainty Expression
\textit{19th-century architecture in the United States is characterized by} -> Does not Contain Uncertainty Expression

As this is an automated metric, we also conduct a human evaluation to validate it. The results can be found in Appendix \ref{App: Human Evaluation}.





\begin{figure}
    \centering
    \includegraphics[width=.9\linewidth]{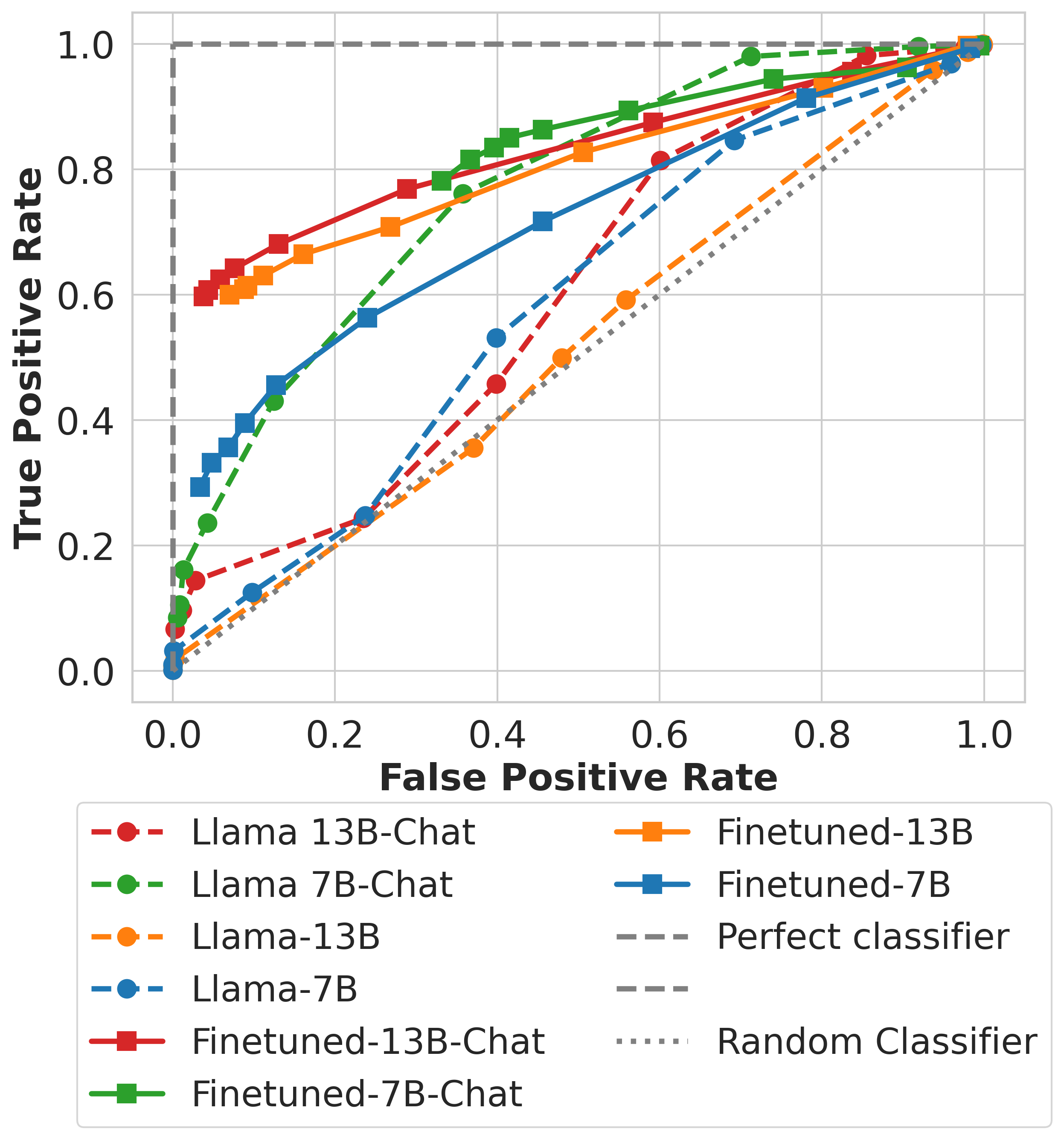}
    \caption{ROC curve comparing fine-tuned and original models in Direct Question-Answering on the KUQ test set (1377 samples)}
    \label{fig:ROC}
\end{figure}

In our evaluations, we utilize two key metrics: the F1 score and the Equal Error Rate (EER). The F1-score, derived from the similarity metric, is calculated with the positive class being either unknown questions or the chosen category. Concurrently, we measure the EER to assess the balance between false acceptance and false rejection rates, offering a holistic view of the system's performance. Additionally, we evaluate answer accuracy on Known Questions, where a response is deemed correct if it includes the ground truth answer. This metric helps us understand potential regressions in known questions that may arise due to fine-tuning.



\section{Experiments}
\label{Section:Experiments}

In this section, we dive into a series of experiments centered around "known-unknown" questions. Our main aim is to see how well current Large Language Models (LLMs) handle these highly uncertain queries. We discuss the experiments and their results in more detail in the following sections.

\subsection{Known vs Unknown}
\label{Section:KnownVsUnknown}


\begin{figure}[]
    \centering
    \includegraphics[width=.9\linewidth]{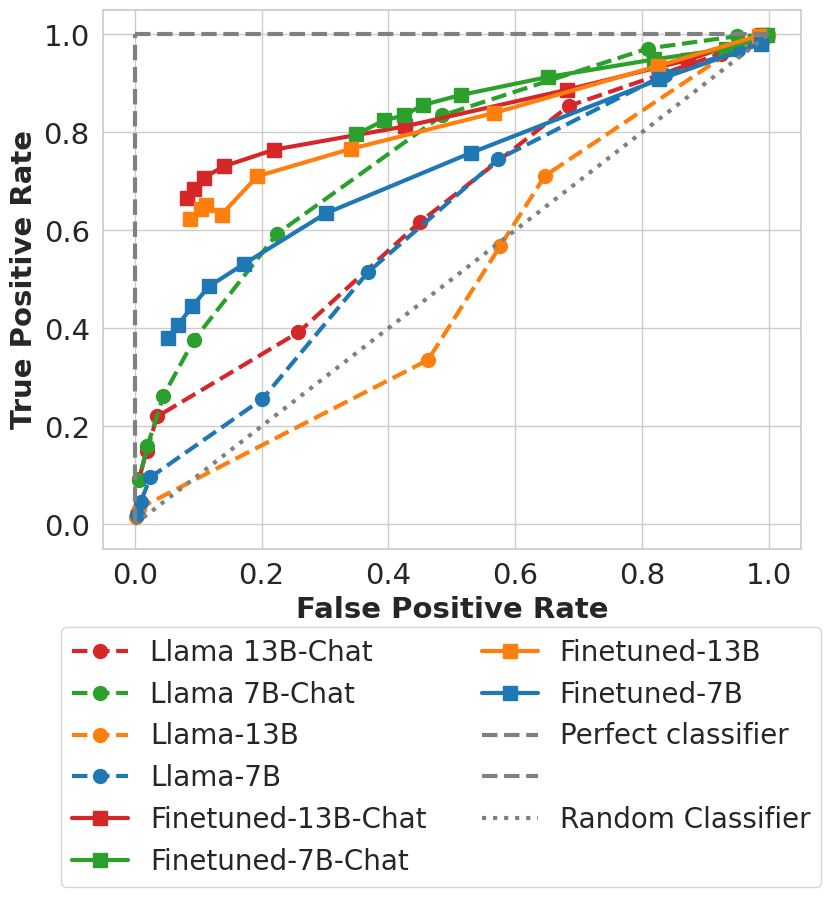}
    \caption{ROC curve comparing the performance of original models and models fine-tuned on KUQ, evaluated on the SelfAware dataset.}
    \label{fig:ROC-selfaware}
\end{figure}

We want to determine the ability of Large Language Models to distinguish between Known and Unknown Questions.
Furthermore, we aim to reproduce the closest to a real-world scenario as possible, where the model may be prompted with questions and needs to provide the answer, with a varying level of uncertainty. 

We present the results on the KUQ evaluation set before fine-tuning for GPT-3.5, GPT-4, and the Llama Models. We also present the results after fine-tuning for Llama-7B, Llama-13B, and its derived chat versions. 

Figure \ref{fig:ROC} presents the ROC curve (receiver operating characteristic curve), showing the performance of the classification at different classification thresholds. From the plot, we extract the EER and the corresponding F1-score. These results are presented in Table \ref{Table:EER Results}. From these results, we can observe there are 2 trends: (1) Bigger models tend to obtain better out-of-the-box results. (2) Chat versions have a better performance in differentiating between known and unknown questions. This may be due to some existing similarity between their RLHF data and this experiment. 

In order to test the fine-tuned models in an out-of-domain distribution, we also show the results after fine-tuning on the KUQ dataset and evaluating the SelfAware \cite{selfaware} dataset. The results are shown in Figure \ref{fig:ROC-selfaware} and Table \ref{Table:EER-Results-selfaware}. We observe fine-tuning on KUQ improves the general ability to differentiate between known and unknown questions. In general, we can see a similar behavior of the models to evaluation on the KUQ dataset. 


\begin{table}[]
\centering
\scalebox{0.9}{
\begin{tabular}{cll}
\hline
\textbf{Model}       & \textbf{EER}     & \textbf{F1}      \\ \hline
\multicolumn{3}{c}{\textit{Closed-source models (OpenAI)}} \\ \hline
GPT-4                & 0.251            & 0.762            \\
GPT-3.5              & 0.271            & 0.747            \\ \hline
\multicolumn{3}{c}{\textit{Original Open-Source Models}}   \\ \hline
Llama 70B-Chat       & 0.318            & 0.721            \\
Llama 13B-Chat       & 0.451            & 0.545            \\
Llama 7B-Chat        & 0.309            & 0.742            \\
Llama-70B            & 0.488            & 0.513            \\
Llama-13B            & 0.489            & 0.512            \\
Llama-7B             & 0.433            & 0.561            \\ \hline
\multicolumn{3}{c}{\textit{Fine-tuned Open-Source Models}}  \\ \hline
Fine-tuned-13B-Chat   & 0.252 \small \textcolor{lightergreen}{$\downarrow$44\%}     & 0.788 \small \textcolor{lightergreen}{$\uparrow$45\%}    \\
Fine-tuned-7B-Chat    & 0.275 \small \textcolor{lightergreen}{$\downarrow$11\%}   & 0.742 \small \textcolor{lighteryellow}{0\%}       \\
Fine-tuned-13B        & 0.284 \small \textcolor{lightergreen}{$\downarrow$42\%}   & 0.735 \small \textcolor{lightergreen}{$\uparrow$43\%}    \\
Fine-tuned-7B         & 0.355 \small \textcolor{lightergreen}{$\downarrow$18\%}   & 0.685 \small \textcolor{lightergreen}{$\uparrow$22\%}    \\ \hline
\end{tabular}
}
\vspace{-8pt}
\caption{Table of Direct Question Answering results on the KUQ Dataset, showing Equal Error Rate (EER) and corresponding F1 scores (lower EER indicates better performance, higher F1 shows better performance).}
\label{Table:EER Results}
\end{table}

\begin{table}[]
\centering
\scalebox{0.9}{
\begin{tabular}{cll}
\hline
\textbf{Model}       & \textbf{EER}     & \textbf{F1}      \\ \hline
\multicolumn{3}{c}{\textit{Original Open-Source Models}}   \\ \hline
Llama 13B-Chat       & 0.419            & 0.469            \\
Llama 7B-Chat        & 0.319            & 0.659            \\
Llama-13B            & 0.529            & 0.362            \\
Llama-7B             & 0.423            & 0.454            \\ \hline
\multicolumn{3}{c}{\textit{(KUQ) Fine-tuned Open-Source Models}}  \\ \hline
Fine-tuned-13B-Chat   & 0.232 \small \textcolor{lightergreen}{$\downarrow$45\%}     & 0.739 \small \textcolor{lightergreen}{$\uparrow$58\%}    \\
Fine-tuned-7B-Chat    & 0.277 \small \textcolor{lightergreen}{$\downarrow$13\%}   & 0.615 \small \textcolor{lighterred}{$\downarrow$7\%}       \\
Fine-tuned-13B        & 0.263 \small \textcolor{lightergreen}{$\downarrow$50\%}   & 0.686 \small \textcolor{lightergreen}{$\uparrow$90\%}    \\
Fine-tuned-7B         & 0.343 \small \textcolor{lightergreen}{$\downarrow$19\%}   & 0.595 \small \textcolor{lightergreen}{$\uparrow$31\%}    \\ \hline
\end{tabular}
}
\vspace{-8pt}
\caption{Table comparing Direct Question Answering results evaluated on the SelfAware Dataset for original and KUQ fine-tuned models, detailing Equal Error Rate (EER) and F1 scores (lower EER signifies better performance, higher F1 indicate better performance).}
\label{Table:EER-Results-selfaware}
\vspace{-15pt}
\end{table}

\subsection{Effects on Fine-tuning}

\begin{figure*}[h]
  \centering

  \begin{subfigure}{0.24\textwidth}
    \includegraphics[width=\linewidth]{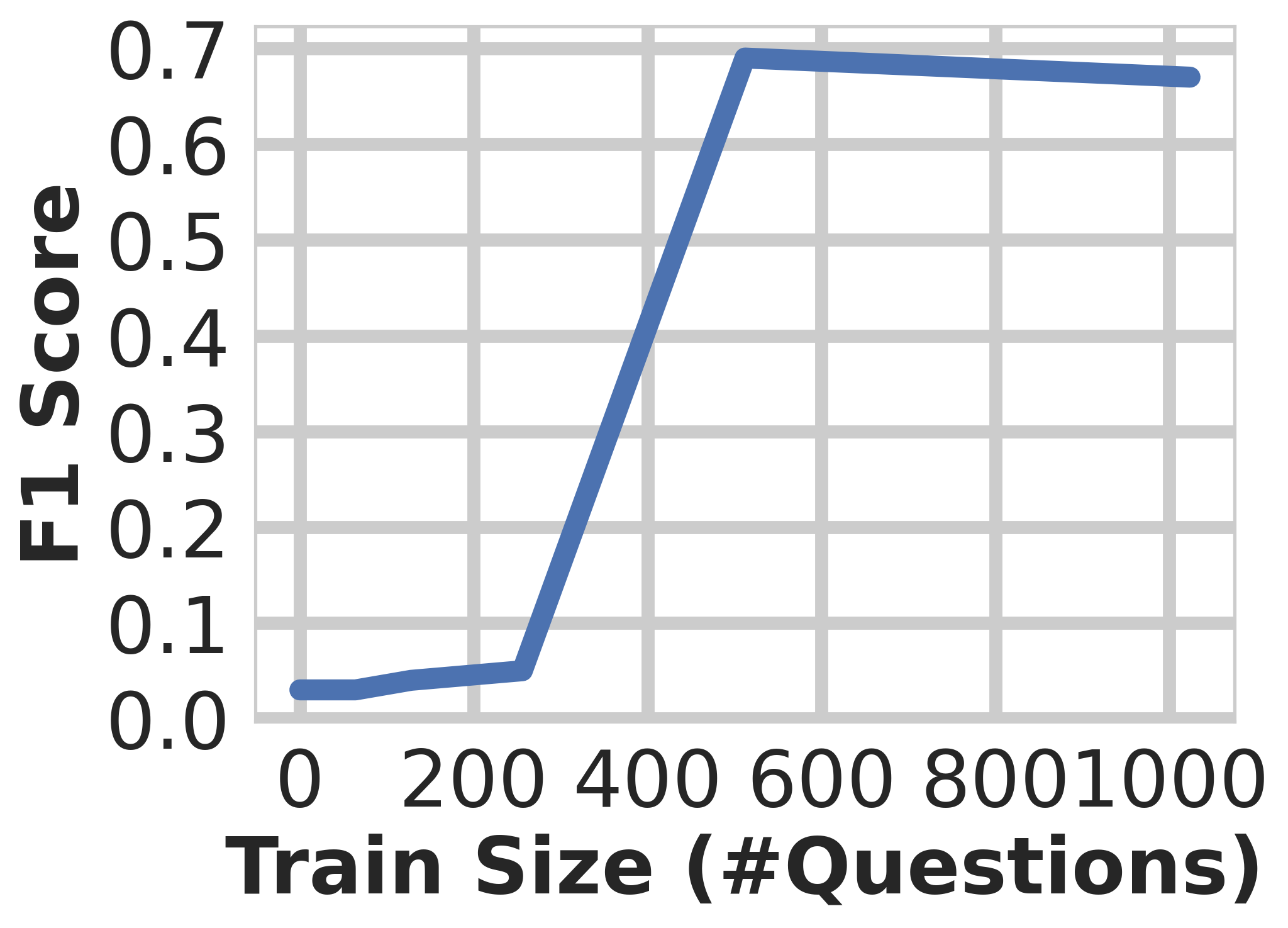}
    \caption{Llama-7B}
  \end{subfigure}
  \hfill
  \begin{subfigure}{0.24\textwidth}
    \includegraphics[width=\linewidth]{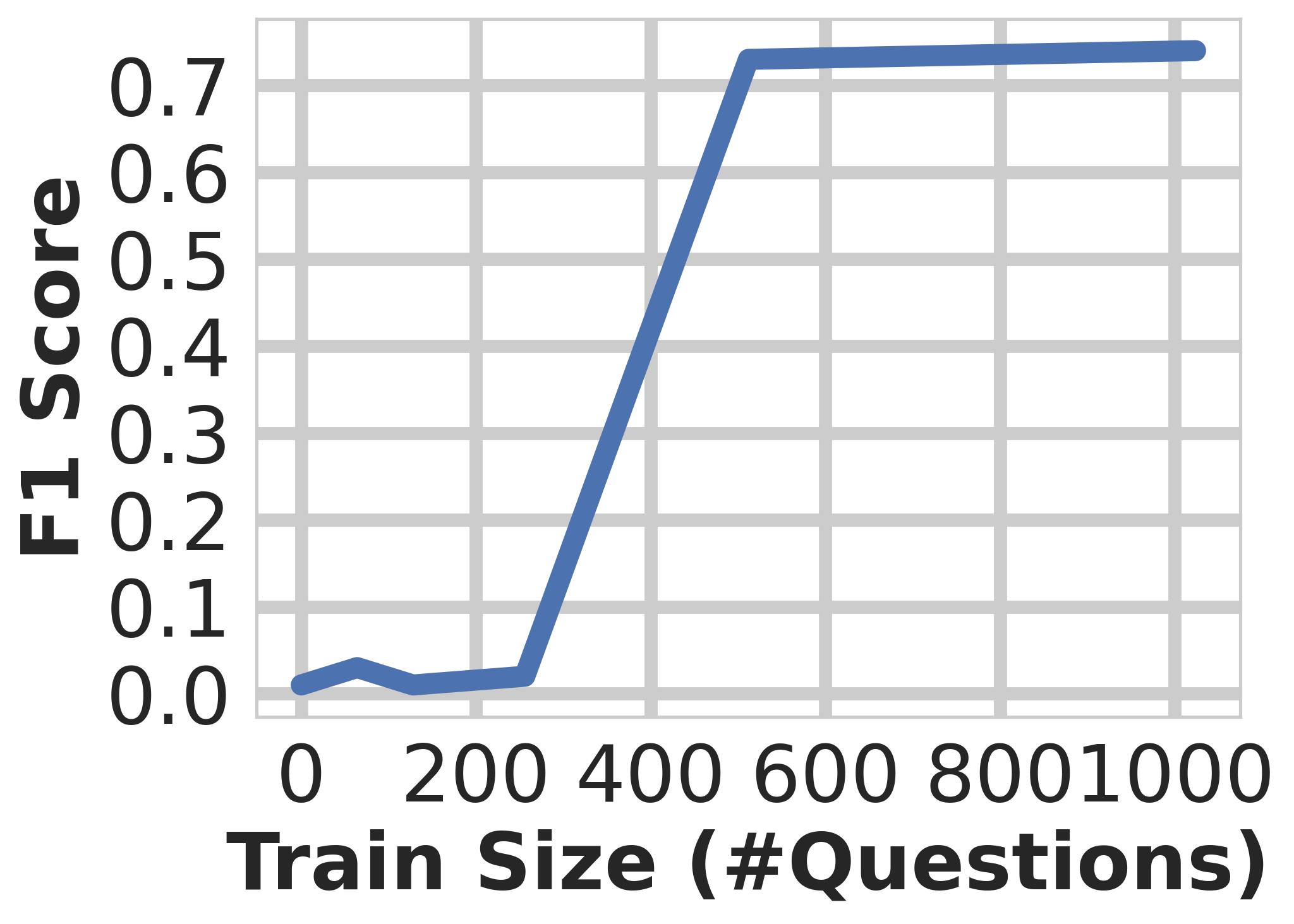}
    \caption{Llama-13B}
  \end{subfigure}
  \hfill
  \begin{subfigure}{0.24\textwidth}
    \includegraphics[width=\linewidth]{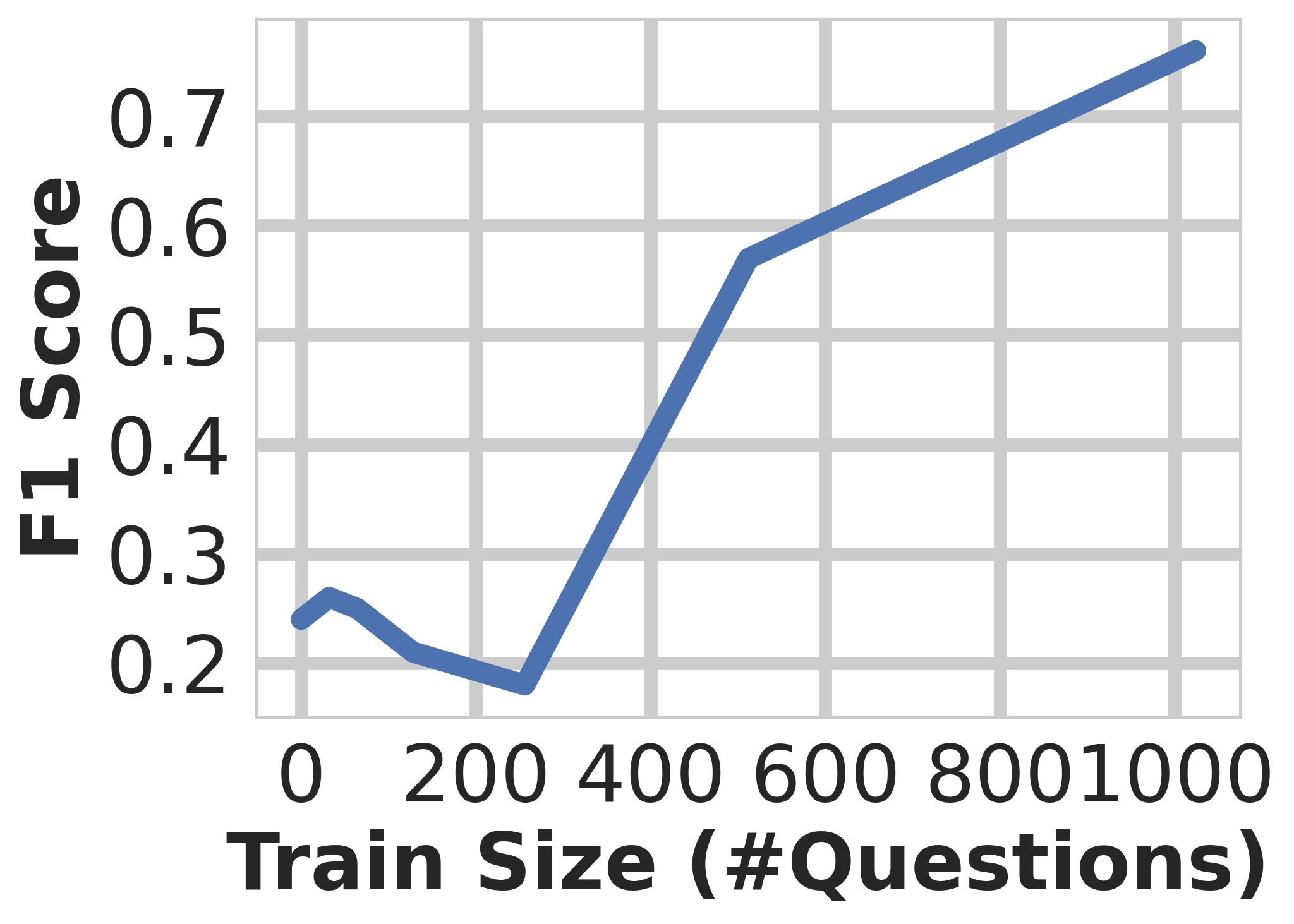}
    \caption{Llama-7B Chat}
  \end{subfigure}
  \hfill
  \begin{subfigure}{0.24\textwidth}
    \includegraphics[width=\linewidth]{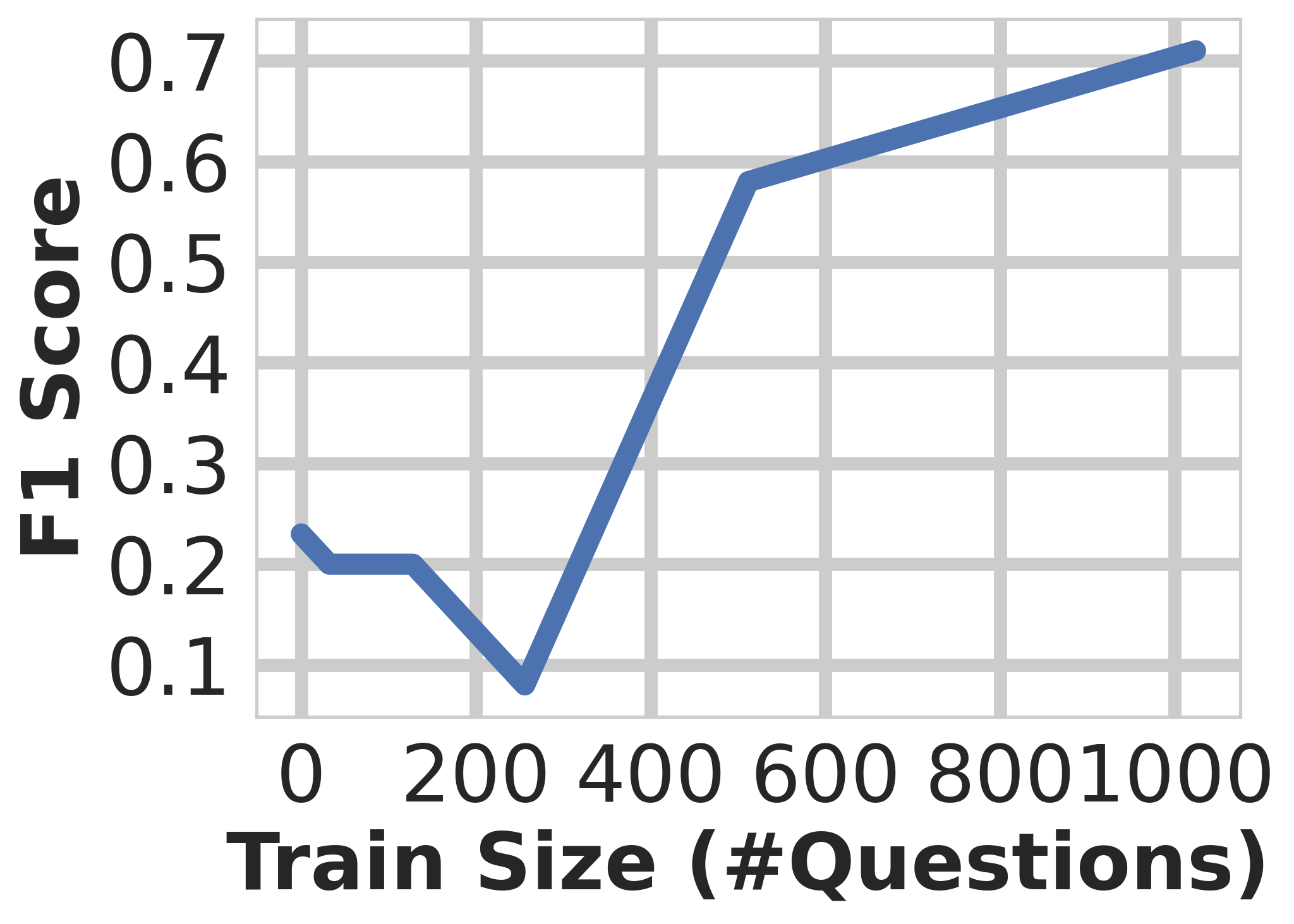}
    \caption{Llama-13B Chat}
  \end{subfigure}

  \caption{Known vs Unknown Classification of Fine-tuned Llama Models on KUQ dataset: Image shows F1-Scores for Known vs Unknown Questions. }
  \label{Fig:f1_finetune_size}
\end{figure*}
\vspace{-4pt}

\begin{figure*}[h]
  \centering

  \begin{subfigure}{0.24\textwidth}
    \includegraphics[width=\linewidth]{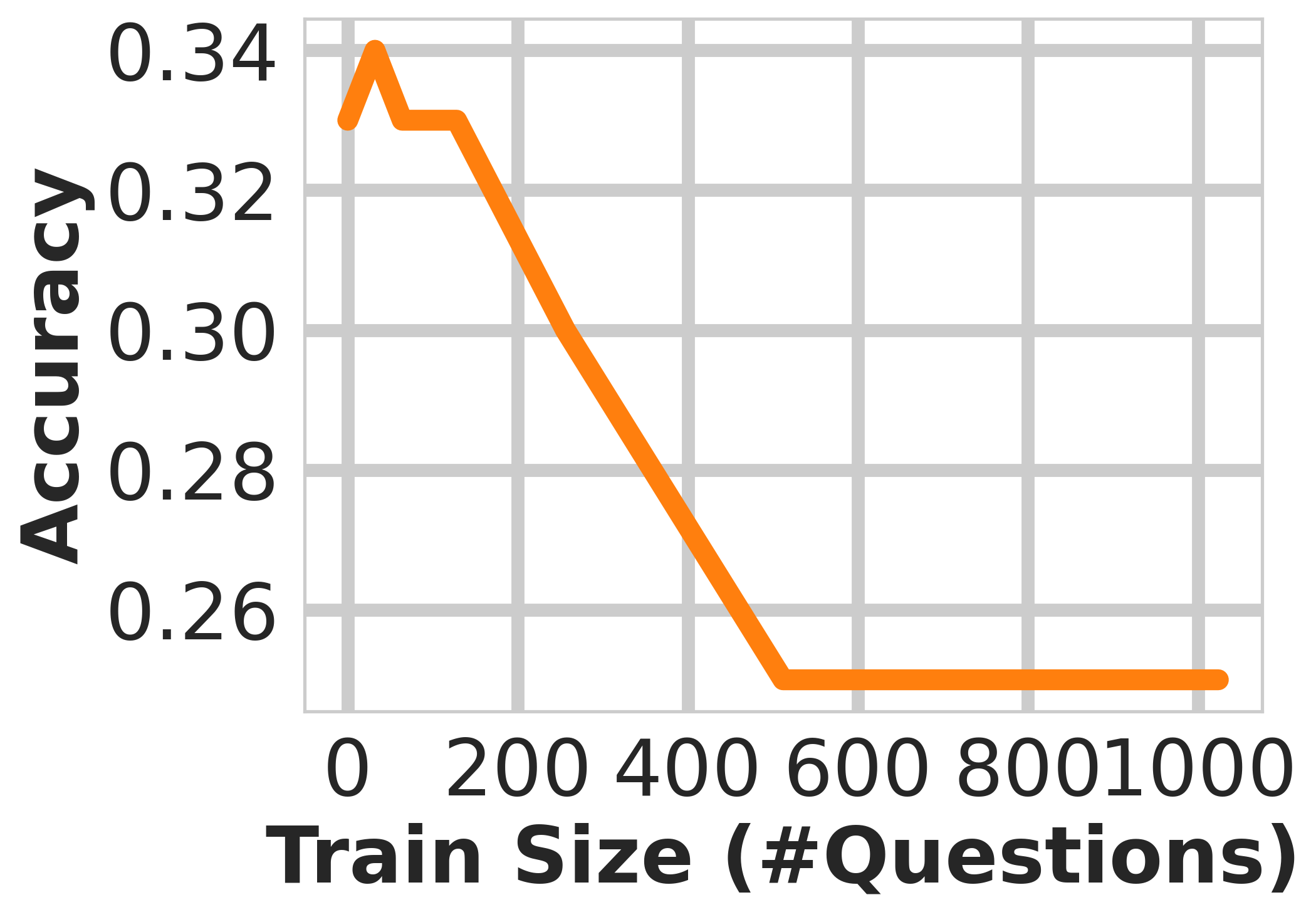}
    \caption{Llama-7B}
  \end{subfigure}
  \hfill
  \begin{subfigure}{0.24\textwidth}
    \includegraphics[width=\linewidth]{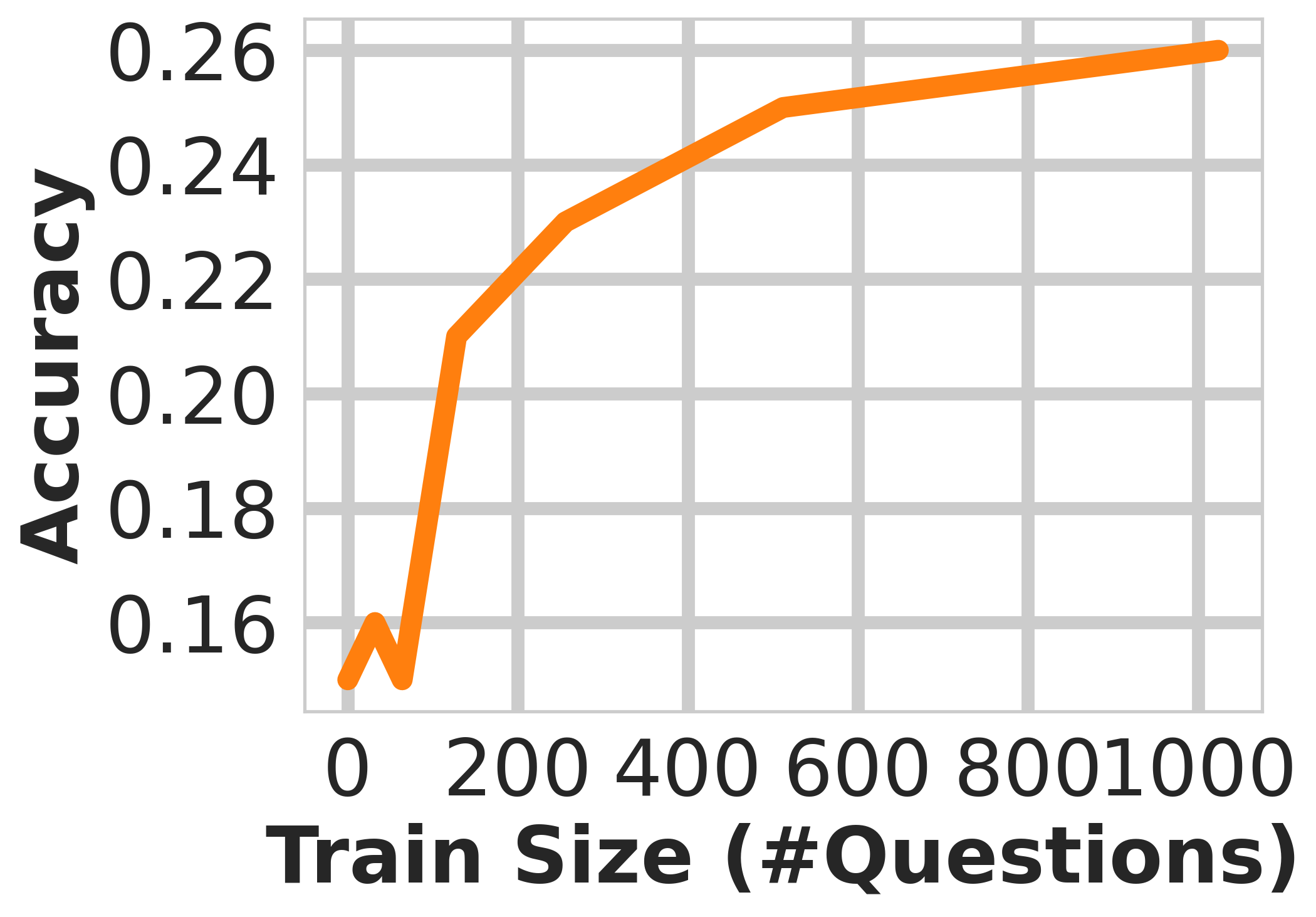}
    \caption{Llama-13B}
  \end{subfigure}
  \hfill
  \begin{subfigure}{0.24\textwidth}
    \includegraphics[width=\linewidth]{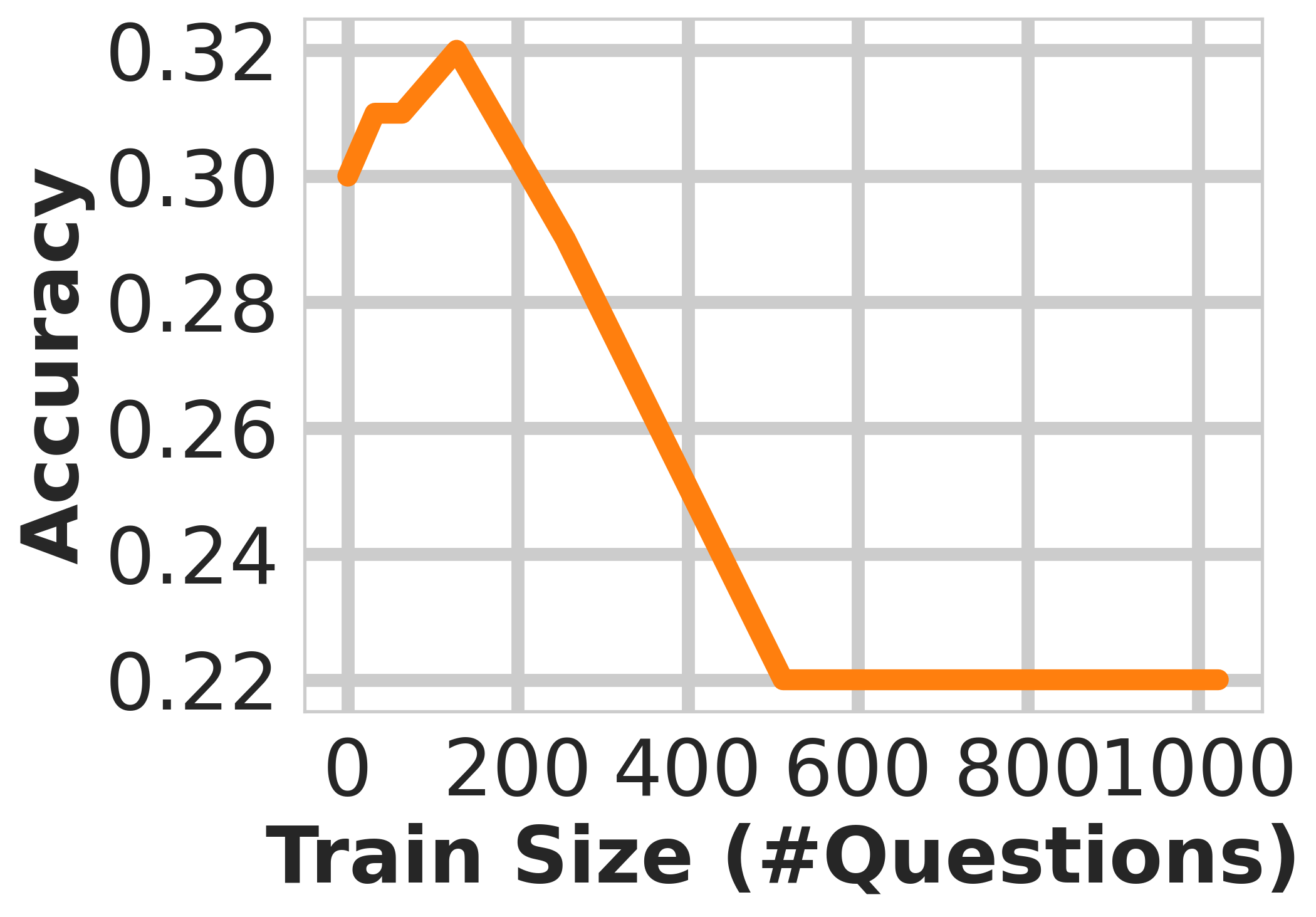}
    \caption{Llama-7B Chat}
  \end{subfigure}
  \hfill
  \begin{subfigure}{0.24\textwidth}
    \includegraphics[width=\linewidth]{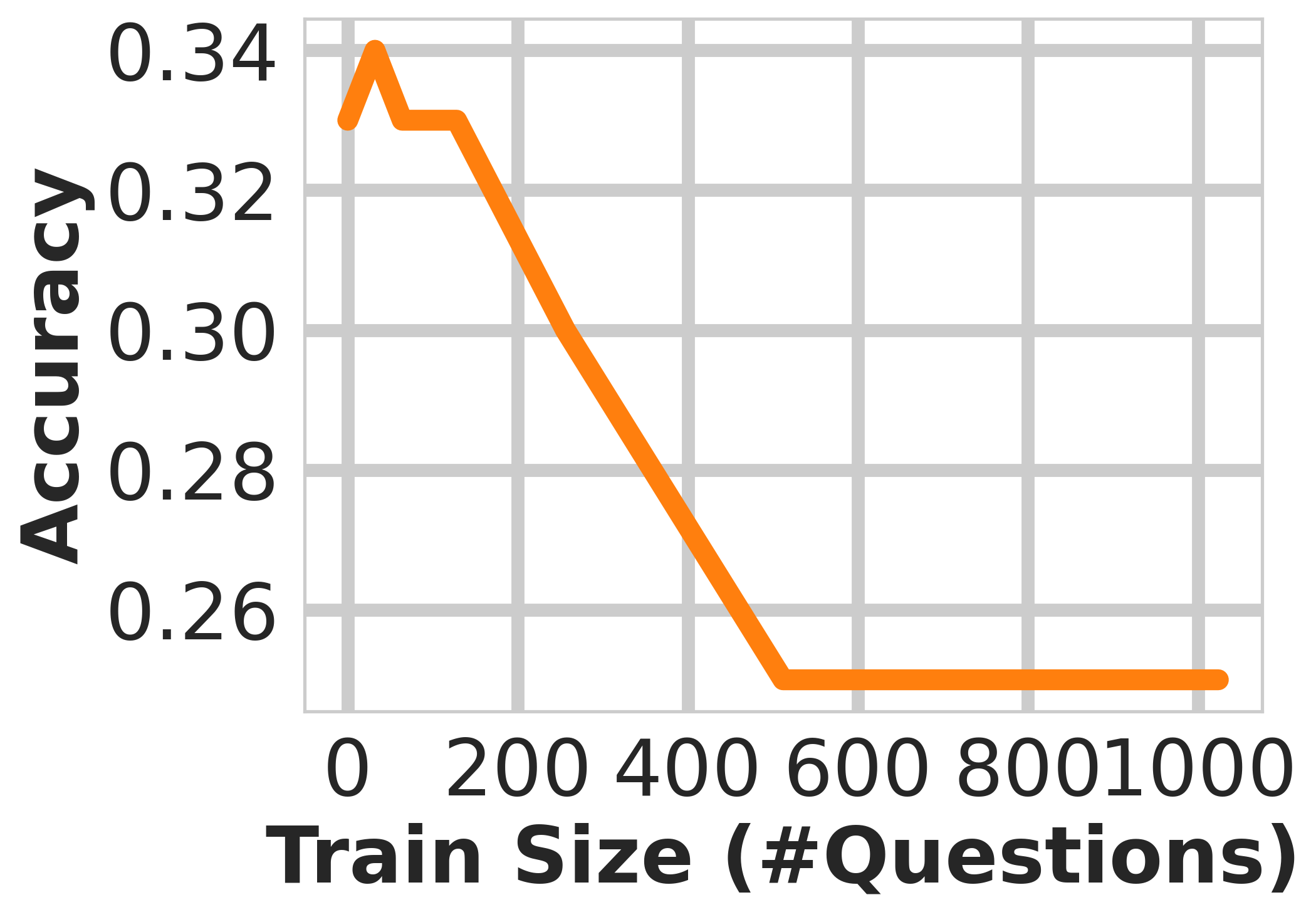}
    \caption{Llama-13B Chat}
  \end{subfigure}

  \caption{Known Questions Answer Accuracy of Fine-tuned Llama Models: This image shows how accuracy to answer correctly changes after fine-tuning on the KUQ dataset. It shows the trade-off between fine-tuning and model knowledge. This shows the minimum number of samples needed to gain some ability to distinguish known vs unknowns.}
  \label{Fig:acc_finetune_size}
  \vspace{-12pt}
\end{figure*}

In this section, we analyze the impact of fine-tuning on model performance,  with a specific emphasis on determining the minimal dataset size required for effectively learning to distinguish between known and unknown questions. 

We specifically investigate the number of training samples necessary for models to discern between known and unknown questions. To this end, we have conducted experiments using the Llama-7b and Llama-13b models, training and evaluating them on datasets ranging from 32 to 1024 questions.

Figure \ref{Fig:f1_finetune_size} presents the F1-scores for the fine-tuned models in distinguishing between known and unknown questions. Our findings indicate that basic models show improved performance at approximately 256 samples, whereas chat-oriented models require around 512 samples. We hypothesize that this difference may stem from the chat models' pre-existing training in question-answering tasks, which could lead to inherently better initial performance but also necessitate more data for significant behavioral adjustments during question-answering.

Additionally, our analysis reveals a trade-off inherent in fine-tuning with the KUQ dataset. While it enhances the model's capability to express uncertainty when confronted with unknown questions, it also results in a slight decrease in overall accuracy in answering the known questions correctly. This effect can be seen in Figure \ref{Fig:acc_finetune_size}, where the accuracy to known questions drops at the same time as the models gain the ability to differentiate known-unknown questions in their responses. Overall, the accuracy of known questions drops slightly in the fine-tuned models, as shown in Appendix \ref{App:Effects on Accuracy}.

\subsection{Downstream Application: MultiAgent Debate}
\label{Section:MultiAgentDebate}


In \cite{multiagent_debate}, they have demonstrated an approach where multiple language model instances propose and debate their responses and reasoning processes over multiple rounds to arrive at a common final answer. Their findings show how this debate improves the results over a single model on several reasoning, factuality, and question-answering tasks.  

A highlighted insight from multiagent debate is the fact that models may converge to a final answer, even when the answer is not correct. Despite arriving at the same answer, models can confidently affirm that their answers are correct. This phenomenon could potentially be attributed to a limitation inherent in the models' design: the inability to accurately represent uncertainty within their response generation process. 

\pagebreak

With the hypothesis that expressing uncertainty can help downstream applications, we want to show that the fine-tuned models on KUQ can potentially better understand question uncertainty and express it accordingly, which leads to a performance increase in a multiagent debate. 

\paragraph{Experiment}

We follow the same approach as the one presented in the original Multiagent Debate paper \cite{multiagent_debate}, where a model is first asked to generate the answer to the proposed questions. In the next turn, the model is presented with the response from the previous turn and asked to improve it. The experiments have been carried out on the default original settings. This means we select the number of agents = 3 and the number of rounds = 2, which provides a good trade-off between results and speed to generate the results. The prompts used in this experiment can be found in Appendix \ref{App:Prompts}.

\paragraph{Data}

For this experiment, we have evaluated the following datasets to test the LLM abilities to reason and provide complex knowledge: 

\noindent \textbullet \hspace{2pt} \texttt{MMLU} \cite{mmlu}, a dataset to measure the factuality of language models in answering questions typically found in different exams. 

\noindent \textbullet \hspace{2pt} \texttt{CommonSenseQA (CSQA)} \cite{commonsenseQA}, a question-answering dataset for commonsense questions based on the knowledge from ConceptNet \cite{concepnet}

\noindent \textbullet \hspace{2pt} \texttt{AI2 Reasoning Challenge (ARC)} \cite{arc}, a question-answering dataset containing natural, grade-school science questions that require knowledge and reasoning abilities.

\noindent \textbullet \hspace{2pt} \texttt{BIG-Bench Chess State Tracking} \cite{bigBench}, a dataset that contains an unfinished sequence of moves from a chess game paired with a set of valid moves to complete the sequence.



\paragraph{Evaluation}

We evaluate our approach to four distinct reasoning tasks, which involve reasoning and extracting factual knowledge from the models. These tasks are presented through three datasets (MMLU, CSQA, ARC), each comprising multiple-choice Question-Answer pairs. In these datasets, only one answer option is correct, and we assess the model's accuracy in choosing this option. Additionally, we analyze the Chess-State Tracking benchmark, where a set of given chess moves is deemed correct, and we measure the model's accuracy in producing any appropriate move from that given list. This research follows the methodology of the original paper, focusing on a subset of the entire dataset. We conducted several experimental runs using different seeds to ensure a representative sample. Our comparison involves the fine-tuned model LLama2-7B-Chat and its predecessor, the original Llama2-7B-Chat.

\paragraph{Results}

\begin{figure}
    \centering
    \includegraphics[width=\linewidth]{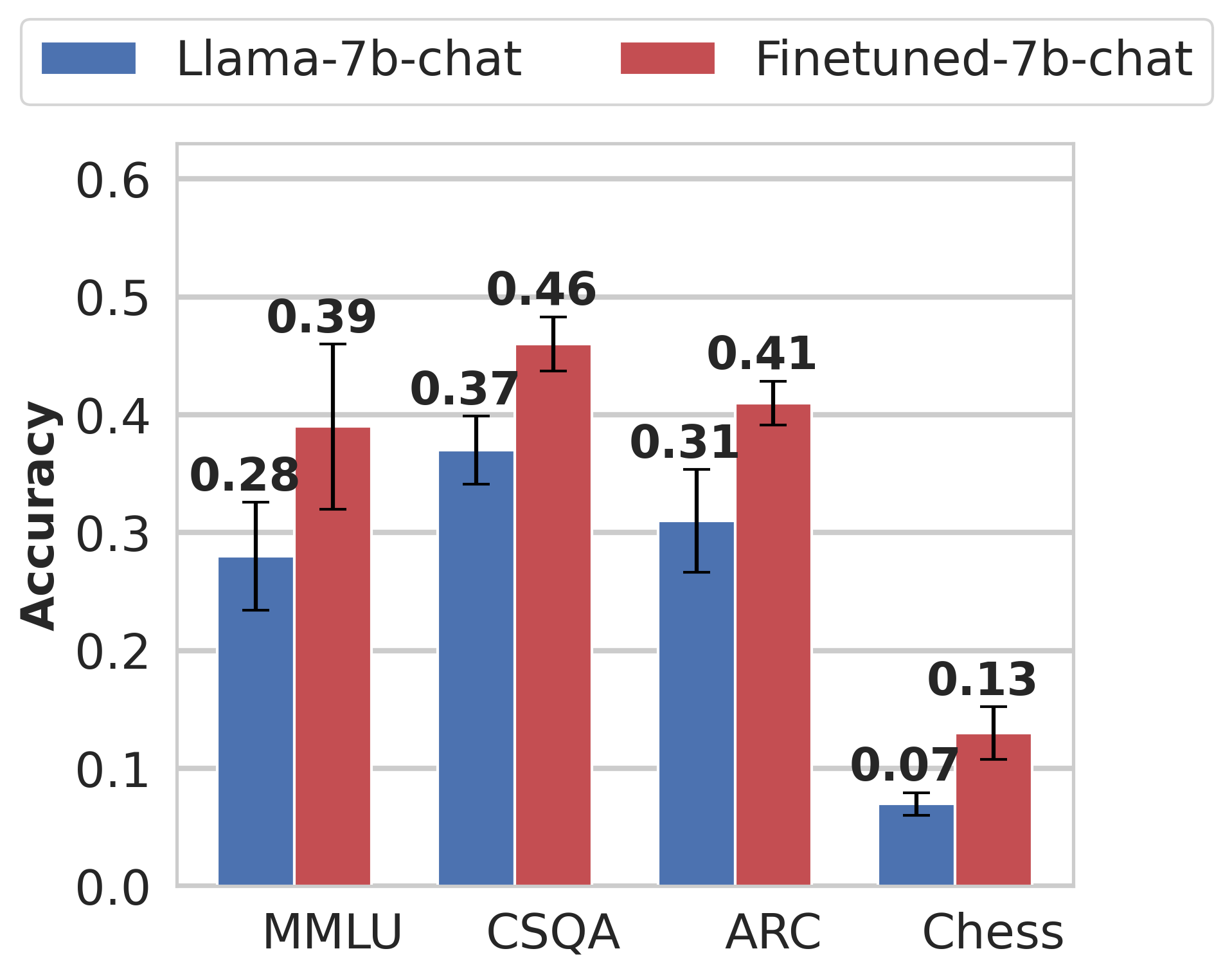}
    \caption{Downstream Application: Multiagent Debate. The figure presents accuracy results from Experiment \cref{Section:MultiAgentDebate} over 4 benchmark datasets for the LLama-7b-chat model and the model fine-tuned on our KUQ dataset. It shows how the fine-tuned model is able to achieve better results due to the expression of uncertainty. }
    \label{fig:multiagent-debate}
    \vspace{-10pt}

\end{figure}

Figure \ref{fig:multiagent-debate} illustrates the performance outcomes of our experiment. It is evident that the fine-tuned model utilizing the KUQ dataset attains a performance enhancement in the multiagent debates when compared to the baseline model (Llama2-7B-Chat). Despite this advancement, a human analysis of the responses denotes a deficiency in the models' ability to consistently represent uncertainty. Further research is necessary to accurately represent the models' inherent uncertainty, in addition to the explicit uncertainty derived from the questions. 



\paragraph{Analysis}

\begin{table}[]
\centering
\scalebox{0.85}{
\begin{tabular}{cll}
\toprule
Models            & \begin{tabular}{@{}c@{}c@{}} Uncertainty \\ Similarity \\ Expressions \end{tabular} & \begin{tabular}{@{}c@{}c@{}} Percentage \\ Selected \\ Expressions \end{tabular} \\
\hline
\multicolumn{3}{c}{\textit{MMLU}}                     \\
\hline
7B-Chat           & 0.515       & 0.07       \\
Fine-tuned 7B-Chat & 0.610  \small \textcolor{lightergreen}{$\uparrow$18\%}      & 0.110  \small \textcolor{lightergreen}{$\uparrow$57\%}      \\
\hline
\multicolumn{3}{c}{\textit{CommonSenseQA}}                     \\
\hline
7B-Chat           & 0.525       & 0.064      \\
Fine-tuned 7B-Chat & 0.631 \small \textcolor{lightergreen}{$\uparrow$20\%}      & 0.137  \small \textcolor{lightergreen}{$\uparrow$114\%}    \\
\hline
\multicolumn{3}{c}{\textit{AI2 Reasoning Challenge}}                      \\
\hline
7B-Chat           & 0.491       & 0.044      \\
Fine-tuned 7B-Chat & 0.608 \small \textcolor{lightergreen}{$\uparrow$24\%}      & 0.083   \small \textcolor{lightergreen}{$\uparrow$88\%}   \\
\hline
\multicolumn{3}{c}{\textit{Chess Validity}}           \\
\hline
7B-Chat           & 0.550       & 0.026      \\
Fine-tuned 7B-Chat & 0.550  \small \textcolor{lighteryellow}{0\%}     & 0.040   \small \textcolor{lightergreen}{$\uparrow$54\%}  \\
\bottomrule
\end{tabular}
}
\vspace{-6pt}
\caption{Analysis of Uncertainty in Debate Responses. This table indicates the similarity between generated sentences and a predefined list of uncertainty expressions, using the SimCSE model \cite{simcse}. It presents the percentage of debate texts closely matching the uncertainty expressions list, highlighting a greater prevalence of uncertainty in responses from the fine-tuned model. Results show a higher number of responses from the fine-tuned model containing uncertainty expression.}
\label{Table:Uncertainty-Analysis}
\end{table}

The results from the experiments show an improved accuracy of the fine-tuned models in the benchmark datasets. In order to validate the hypothesis that uncertainty expression helped increase the accuracy, we measure the uncertainty expressions generated in the responses. To do so, we compare the generated texts to the list of uncertainty expressions from Appendix \ref{App:List of Uncertainty Expressions} and report the average similarity according to SimCSE \cite{simcse}. We also report the average number of responses with a high similarity -- higher than 0.75 --. The analysis for all 4 datasets is presented in Table \ref{Table:Uncertainty-Analysis}. In all cases, the fine-tuned models express more uncertainty over the questions. This may be due to the fact that some questions can be interpreted ambiguously, as seen in an example debate response in Appendix \ref{App:Examples of Uncertainty Improving Debate Responses}.


\section{Conclusion}

This work explores how open-source LLMs handle Known-Unknown questions, which are characterized by high uncertainty and the expectation of non-confident answers. We introduce a new dataset with \textbf{K}nown and \textbf{U}nknown \textbf{Q}uestions, \textbf{KUQ}. In addition, a categorization of unknown questions is introduced, offering different reasons for them being unknown. 

Along this work, we evaluate the current open-source models in open-ended question-answering on our KUQ dataset. We evaluate (1) the ability to tell the difference between known and unknown questions and (2) the ability to distinguish between the different categories of questions. 

Finally, we show how the expression of uncertainty may help in specific applications of Large Language Models. In particular, we show how the fine-tuned model on KUQ improves the results of multiagent debate when compared to the baseline original model.  

Future research directions should focus on enhancing evaluation techniques and tackling the challenge of gauging model epistemic uncertainty, potentially leading to broader applications. Investigating the capacity of LLMs to convey their uncertainty probabilities is also a key area of interest.

\section*{Limitations}

This paper acknowledges several limitations encountered during its research process.

Initially, the task of generating known-unknown questions presents inherent complexity. While these questions have been validated as known-unknowns through human assessment, there remains a possibility of contention regarding their categorization. The compilation of known questions datasets was curated to include a broad set of questions and topics.

Moreover, the methodology for evaluating open-ended question-answering tasks continues to be a subject of ongoing discourse and investigation within the academic community. In this study, we employed a similarity metric to measure uncertainty expressions, acknowledging that the chosen similarity threshold is a variable factor influencing the results. This approach is consistent with methodologies employed in prior research.

Lastly, the decision to fine-tune and evaluate the Llama 2 models was influenced by their significance and popularity in the open-source community at the time of this study. The choice of smaller models (7B, 13B) over larger variants (70B) was dictated by the computational resources available during the research period.

\section*{Ethics Statement}

Human evaluation was conducted through crowdsourcing using the Amazon Mechanical Turk platform. To ensure the quality of our experiments, we only considered workers with a HIT approval rating of at least 95\% from the Mechanical Turk Masters pool. We compensated the workers at a rate of \$0.25 per task. We estimate each task can be completed in 1 minute or less and therefore it translates to a rate of \$15.0 per hour, which exceeds the federal minimum wage in the USA during the time of our research. The data annotation is classified as an approved exempt protocol from the IRB. Details about the tasks, including screenshots and task descriptions of each Mechanical Turk study are included in the appendices sections.

\section*{Acknowledgements}

This work was supported by the National Science Foundation award \#2048122. The views expressed are those of the authors and do not reflect the official policy or position of the US government.

\bibliography{anthology,custom}

\begin{thebibliography}{31}
\expandafter\ifx\csname natexlab\endcsname\relax\def\natexlab#1{#1}\fi

\bibitem[{Chen et~al.(2021)Chen, Tworek, Jun, Yuan, Pinto, Kaplan, Edwards, Burda, Joseph, Brockman et~al.}]{evaluating_llm_code}
Mark Chen, Jerry Tworek, Heewoo Jun, Qiming Yuan, Henrique Ponde de~Oliveira Pinto, Jared Kaplan, Harri Edwards, Yuri Burda, Nicholas Joseph, Greg Brockman, et~al. 2021.
\newblock \href {https://arxiv.org/abs/2107.03374} {Evaluating large language models trained on code}.
\newblock \emph{ArXiv preprint}, abs/2107.03374.

\bibitem[{Clark et~al.(2018)Clark, Cowhey, Etzioni, Khot, Sabharwal, Schoenick, and Tafjord}]{arc}
Peter Clark, Isaac Cowhey, Oren Etzioni, Tushar Khot, Ashish Sabharwal, Carissa Schoenick, and Oyvind Tafjord. 2018.
\newblock Think you have solved question answering? try arc, the ai2 reasoning challenge.
\newblock \emph{arXiv preprint arXiv:1803.05457}.

\bibitem[{Cole et~al.(2023)Cole, Zhang, Gillick, Eisenschlos, Dhingra, and Eisenstein}]{ambiguousQuestions}
Jeremy~R Cole, Michael~JQ Zhang, Daniel Gillick, Julian~Martin Eisenschlos, Bhuwan Dhingra, and Jacob Eisenstein. 2023.
\newblock \href {https://arxiv.org/abs/2305.14613} {Selectively answering ambiguous questions}.
\newblock \emph{ArXiv preprint}, abs/2305.14613.

\bibitem[{Du et~al.(2023)Du, Li, Torralba, Tenenbaum, and Mordatch}]{multiagent_debate}
Yilun Du, Shuang Li, Antonio Torralba, Joshua~B Tenenbaum, and Igor Mordatch. 2023.
\newblock Improving factuality and reasoning in language models through multiagent debate.
\newblock \emph{arXiv preprint arXiv:2305.14325}.

\bibitem[{Farkas et~al.(2010)Farkas, Vincze, M{\'o}ra, Csirik, and Szarvas}]{uncertainty_hedges}
Rich{\'a}rd Farkas, Veronika Vincze, Gy{\"o}rgy M{\'o}ra, J{\'a}nos Csirik, and Gy{\"o}rgy Szarvas. 2010.
\newblock \href {https://aclanthology.org/W10-3001} {The {C}o{NLL}-2010 shared task: Learning to detect hedges and their scope in natural language text}.
\newblock In \emph{Proceedings of the Fourteenth Conference on Computational Natural Language Learning {--} Shared Task}, pages 1--12, Uppsala, Sweden. Association for Computational Linguistics.

\bibitem[{Gao et~al.(2021)Gao, Yao, and Chen}]{simcse}
Tianyu Gao, Xingcheng Yao, and Danqi Chen. 2021.
\newblock Simcse: Simple contrastive learning of sentence embeddings.
\newblock \emph{arXiv preprint arXiv:2104.08821}.

\bibitem[{Garner and Alexander(1989)}]{metacognition}
Ruth Garner and Patricia~A Alexander. 1989.
\newblock Metacognition: Answered and unanswered questions.
\newblock \emph{Educational psychologist}, 24(2):143--158.

\bibitem[{Hendrycks et~al.(2020)Hendrycks, Burns, Basart, Zou, Mazeika, Song, and Steinhardt}]{mmlu}
Dan Hendrycks, Collin Burns, Steven Basart, Andy Zou, Mantas Mazeika, Dawn Song, and Jacob Steinhardt. 2020.
\newblock Measuring massive multitask language understanding.
\newblock \emph{arXiv preprint arXiv:2009.03300}.

\bibitem[{Hu et~al.(2021)Hu, Shen, Wallis, Allen-Zhu, Li, Wang, Wang, and Chen}]{lora}
Edward~J Hu, Yelong Shen, Phillip Wallis, Zeyuan Allen-Zhu, Yuanzhi Li, Shean Wang, Lu~Wang, and Weizhu Chen. 2021.
\newblock Lora: Low-rank adaptation of large language models.
\newblock \emph{arXiv preprint arXiv:2106.09685}.

\bibitem[{Hu et~al.(2023)Hu, Zhang, Zhao, Huang, and Wu}]{uncertatintyNLP_review}
Mengting Hu, Zhen Zhang, Shiwan Zhao, Minlie Huang, and Bingzhe Wu. 2023.
\newblock \href {https://arxiv.org/abs/2306.04459} {Uncertainty in natural language processing: Sources, quantification, and applications}.
\newblock \emph{ArXiv preprint}, abs/2306.04459.

\bibitem[{Jiang et~al.(2021)Jiang, Araki, Ding, and Neubig}]{when_lms_know}
Zhengbao Jiang, Jun Araki, Haibo Ding, and Graham Neubig. 2021.
\newblock \href {https://doi.org/10.1162/tacl_a_00407} {How can we know when language models know? on the calibration of language models for question answering}.
\newblock \emph{Transactions of the Association for Computational Linguistics}, 9:962--977.

\bibitem[{Jiang et~al.(2020)Jiang, Xu, Araki, and Neubig}]{what_lms_know}
Zhengbao Jiang, Frank~F. Xu, Jun Araki, and Graham Neubig. 2020.
\newblock \href {https://doi.org/10.1162/tacl_a_00324} {How can we know what language models know?}
\newblock \emph{Transactions of the Association for Computational Linguistics}, 8:423--438.

\bibitem[{Joshi et~al.(2017)Joshi, Choi, Weld, and Zettlemoyer}]{triviaqa}
Mandar Joshi, Eunsol Choi, Daniel~S Weld, and Luke Zettlemoyer. 2017.
\newblock Triviaqa: A large scale distantly supervised challenge dataset for reading comprehension.
\newblock \emph{arXiv preprint arXiv:1705.03551}.

\bibitem[{Kadavath et~al.(2022)Kadavath, Conerly, Askell, Henighan, Drain, Perez, Schiefer, Dodds, DasSarma, Tran-Johnson et~al.}]{llm_mostly_know}
Saurav Kadavath, Tom Conerly, Amanda Askell, Tom Henighan, Dawn Drain, Ethan Perez, Nicholas Schiefer, Zac~Hatfield Dodds, Nova DasSarma, Eli Tran-Johnson, et~al. 2022.
\newblock \href {https://arxiv.org/abs/2207.05221} {Language models (mostly) know what they know}.
\newblock \emph{ArXiv preprint}, abs/2207.05221.

\bibitem[{Kuhn et~al.(2023)Kuhn, Gal, and Farquhar}]{semantic_uncertainty}
Lorenz Kuhn, Yarin Gal, and Sebastian Farquhar. 2023.
\newblock \href {https://arxiv.org/abs/2302.09664} {Semantic uncertainty: Linguistic invariances for uncertainty estimation in natural language generation}.
\newblock \emph{ArXiv preprint}, abs/2302.09664.

\bibitem[{Lin et~al.(2022)Lin, Hilton, and Evans}]{model_uncertainty_in_words}
Stephanie Lin, Jacob Hilton, and Owain Evans. 2022.
\newblock \href {https://arxiv.org/abs/2205.14334} {Teaching models to express their uncertainty in words}.
\newblock \emph{ArXiv preprint}, abs/2205.14334.

\bibitem[{Min et~al.(2020)Min, Michael, Hajishirzi, and Zettlemoyer}]{ambigQA}
Sewon Min, Julian Michael, Hannaneh Hajishirzi, and Luke Zettlemoyer. 2020.
\newblock \href {https://doi.org/10.18653/v1/2020.emnlp-main.466} {{A}mbig{QA}: Answering ambiguous open-domain questions}.
\newblock In \emph{Proceedings of the 2020 Conference on Empirical Methods in Natural Language Processing (EMNLP)}, pages 5783--5797, Online. Association for Computational Linguistics.

\bibitem[{Rajpurkar et~al.(2018)Rajpurkar, Jia, and Liang}]{squad2}
Pranav Rajpurkar, Robin Jia, and Percy Liang. 2018.
\newblock \href {https://doi.org/10.18653/v1/P18-2124} {Know what you don{'}t know: Unanswerable questions for {SQ}u{AD}}.
\newblock In \emph{Proceedings of the 56th Annual Meeting of the Association for Computational Linguistics (Volume 2: Short Papers)}, pages 784--789, Melbourne, Australia. Association for Computational Linguistics.

\bibitem[{Rajpurkar et~al.(2016)Rajpurkar, Zhang, Lopyrev, and Liang}]{squad}
Pranav Rajpurkar, Jian Zhang, Konstantin Lopyrev, and Percy Liang. 2016.
\newblock Squad: 100,000+ questions for machine comprehension of text.
\newblock \emph{arXiv preprint arXiv:1606.05250}.

\bibitem[{Sharifi~Noorian et~al.(2022)Sharifi~Noorian, Qiu, Gadiraju, Yang, and Bozzon}]{unknowns_computerVision}
Shahin Sharifi~Noorian, Sihang Qiu, Ujwal Gadiraju, Jie Yang, and Alessandro Bozzon. 2022.
\newblock What should you know? a human-in-the-loop approach to unknown unknowns characterization in image recognition.
\newblock In \emph{Proceedings of the ACM Web Conference 2022}, pages 882--892.

\bibitem[{Speer et~al.(2017)Speer, Chin, and Havasi}]{concepnet}
Robyn Speer, Joshua Chin, and Catherine Havasi. 2017.
\newblock Conceptnet 5.5: An open multilingual graph of general knowledge.
\newblock In \emph{Proceedings of the AAAI conference on artificial intelligence}, volume~31.

\bibitem[{Srivastava et~al.(2022)Srivastava, Rastogi, Rao, Shoeb, Abid, Fisch, Brown, Santoro, Gupta, Garriga-Alonso et~al.}]{bigBench}
Aarohi Srivastava, Abhinav Rastogi, Abhishek Rao, Abu Awal~Md Shoeb, Abubakar Abid, Adam Fisch, Adam~R Brown, Adam Santoro, Aditya Gupta, Adri{\`a} Garriga-Alonso, et~al. 2022.
\newblock \href {https://arxiv.org/abs/2206.04615} {Beyond the imitation game: Quantifying and extrapolating the capabilities of language models}.
\newblock \emph{ArXiv preprint}, abs/2206.04615.

\bibitem[{Szarvas et~al.(2012)Szarvas, Vincze, Farkas, M{\'o}ra, and Gurevych}]{cross_semantic_uncertainty}
Gy{\"o}rgy Szarvas, Veronika Vincze, Rich{\'a}rd Farkas, Gy{\"o}rgy M{\'o}ra, and Iryna Gurevych. 2012.
\newblock \href {https://doi.org/10.1162/COLI_a_00098} {Cross-genre and cross-domain detection of semantic uncertainty}.
\newblock \emph{Computational Linguistics}, 38(2):335--367.

\bibitem[{Talmor et~al.(2018)Talmor, Herzig, Lourie, and Berant}]{commonsenseQA}
Alon Talmor, Jonathan Herzig, Nicholas Lourie, and Jonathan Berant. 2018.
\newblock Commonsenseqa: A question answering challenge targeting commonsense knowledge.
\newblock \emph{arXiv preprint arXiv:1811.00937}.

\bibitem[{Touvron et~al.(2023)Touvron, Martin, Stone, Albert, Almahairi, Babaei, Bashlykov, Batra, Bhargava, Bhosale et~al.}]{llama2}
Hugo Touvron, Louis Martin, Kevin Stone, Peter Albert, Amjad Almahairi, Yasmine Babaei, Nikolay Bashlykov, Soumya Batra, Prajjwal Bhargava, Shruti Bhosale, et~al. 2023.
\newblock Llama 2: Open foundation and fine-tuned chat models.
\newblock \emph{arXiv preprint arXiv:2307.09288}.

\bibitem[{Wang et~al.(2023)Wang, Zhu, and Yang}]{controversialQA}
Zhen Wang, Peide Zhu, and Jie Yang. 2023.
\newblock \href {https://arxiv.org/abs/2302.05061} {Controversialqa: Exploring controversy in question answering}.
\newblock \emph{ArXiv preprint}, abs/2302.05061.

\bibitem[{Wei et~al.(2022)Wei, Wang, Schuurmans, Bosma, Chi, Le, and Zhou}]{chainOfThought}
Jason Wei, Xuezhi Wang, Dale Schuurmans, Maarten Bosma, Ed~Chi, Quoc Le, and Denny Zhou. 2022.
\newblock \href {https://arxiv.org/abs/2201.11903} {Chain of thought prompting elicits reasoning in large language models}.
\newblock \emph{ArXiv preprint}, abs/2201.11903.

\bibitem[{Xiao et~al.(2022)Xiao, Liang, Bhatt, Neiswanger, Salakhutdinov, and Morency}]{uncertaintyQuantification}
Yuxin Xiao, Paul~Pu Liang, Umang Bhatt, Willie Neiswanger, Ruslan Salakhutdinov, and Louis-Philippe Morency. 2022.
\newblock \href {https://aclanthology.org/2022.findings-emnlp.538} {Uncertainty quantification with pre-trained language models: A large-scale empirical analysis}.
\newblock In \emph{Findings of the Association for Computational Linguistics: EMNLP 2022}, pages 7273--7284, Abu Dhabi, United Arab Emirates. Association for Computational Linguistics.

\bibitem[{Yang et~al.(2018)Yang, Qi, Zhang, Bengio, Cohen, Salakhutdinov, and Manning}]{hotpotqa}
Zhilin Yang, Peng Qi, Saizheng Zhang, Yoshua Bengio, William~W Cohen, Ruslan Salakhutdinov, and Christopher~D Manning. 2018.
\newblock Hotpotqa: A dataset for diverse, explainable multi-hop question answering.
\newblock \emph{arXiv preprint arXiv:1809.09600}.

\bibitem[{Yin et~al.(2023)Yin, Sun, Guo, Wu, Qiu, and Huang}]{selfaware}
Zhangyue Yin, Qiushi Sun, Qipeng Guo, Jiawen Wu, Xipeng Qiu, and Xuanjing Huang. 2023.
\newblock \href {https://arxiv.org/abs/2305.18153} {Do large language models know what they don't know?}
\newblock In \emph{Findings of Association for Computational Linguistics (ACL)}.

\bibitem[{Zhou et~al.(2023)Zhou, Jurafsky, and Hashimoto}]{navigating_the_grey_area}
Kaitlyn Zhou, Dan Jurafsky, and Tatsunori Hashimoto. 2023.
\newblock \href {https://arxiv.org/abs/2302.13439} {Navigating the grey area: Expressions of overconfidence and uncertainty in language models}.
\newblock \emph{ArXiv preprint}, abs/2302.13439.

\end{thebibliography}
\bibliographystyle{acl_natbib}

\appendix
\onecolumn 

\section{Dataset Statistics}
\label{appendix:dataset_statistics}

\begin{figure}[h]
    \centering
    \includegraphics[width=\linewidth]{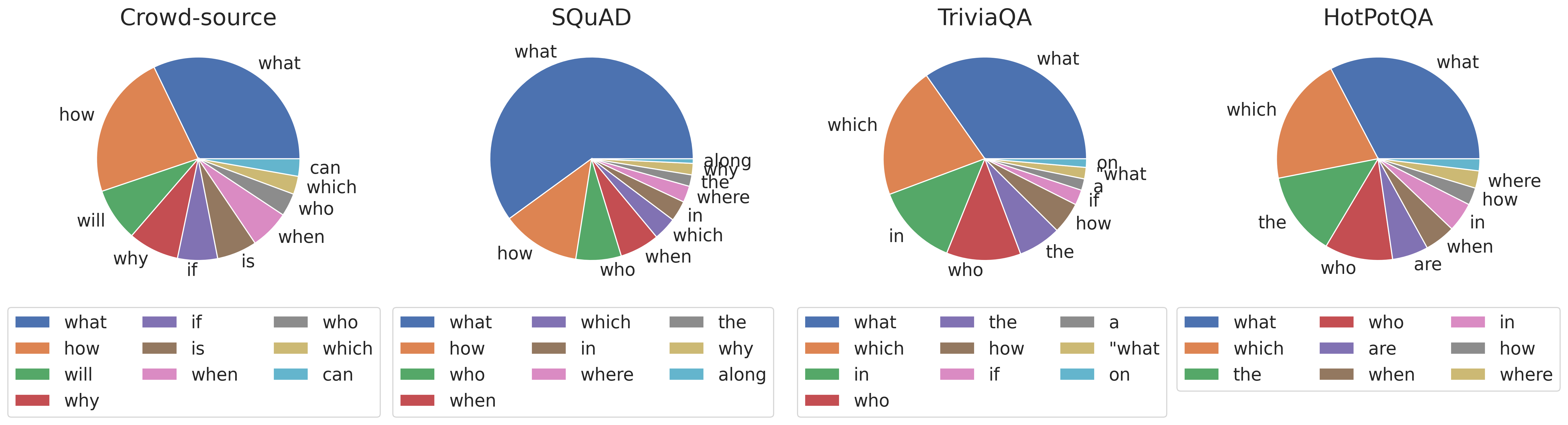}
    \caption{First-word distribution for each data source in the introduced KUQ dataset. }
    \label{fig:start_words}
\end{figure}

This appendix presents, in Figure \ref{fig:start_words}, the First-Word Distribution for each of the sources employed in our Known-Unknown Questions (KUQ) dataset.

\section{Crowd-Source Question Generation and Classification}
\label{App:Crowd-source-Dataset-Generation}

\begin{figure}[h]
  \centering
  \begin{subfigure}[b]{0.58\textwidth}
    \centering
    \includegraphics[width=\textwidth]{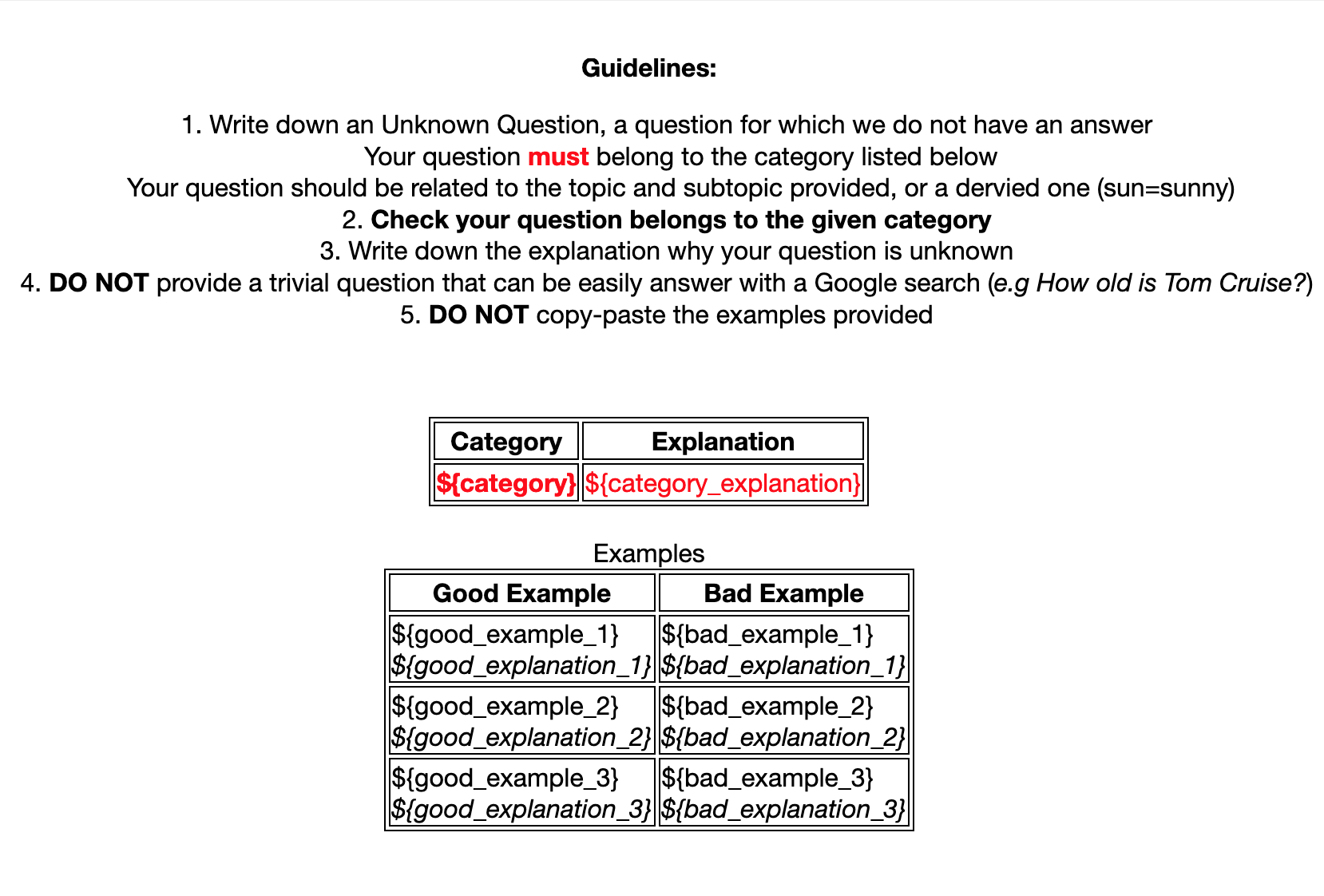}
  \end{subfigure}
  \hfill
  \begin{subfigure}[b]{.61\textwidth}
    \centering
    \includegraphics[width=\textwidth]{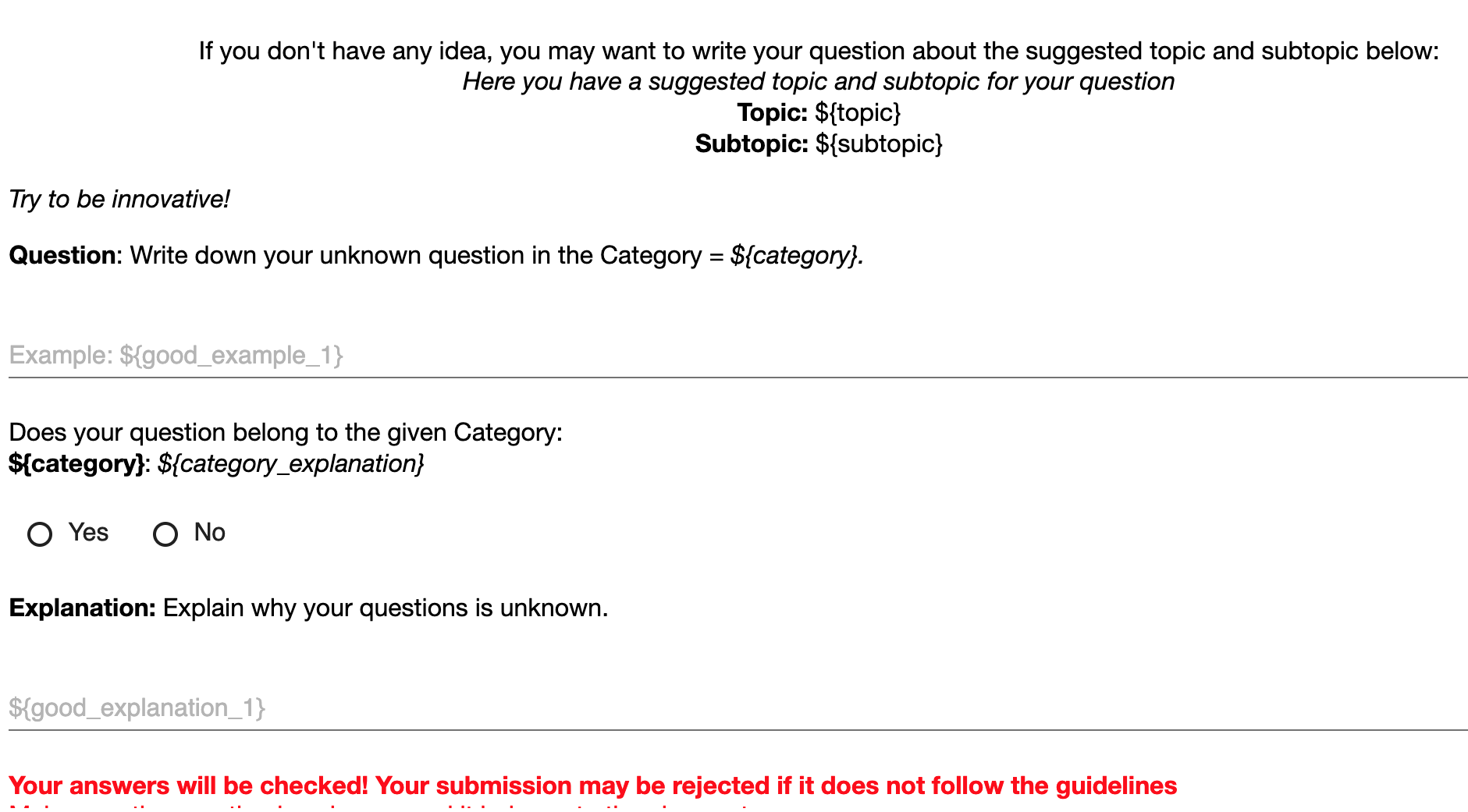}
  \end{subfigure}
  \caption{Crowd-Source Question Generation Screenshot. It shows the guidelines on the left, and the user input screen on the right.}
  \label{fig:turk_question_generation}
\end{figure}

\begin{figure}[htbp]
  \centering
  \begin{subfigure}[b]{0.66\textwidth}
    \centering
    \includegraphics[width=\textwidth]{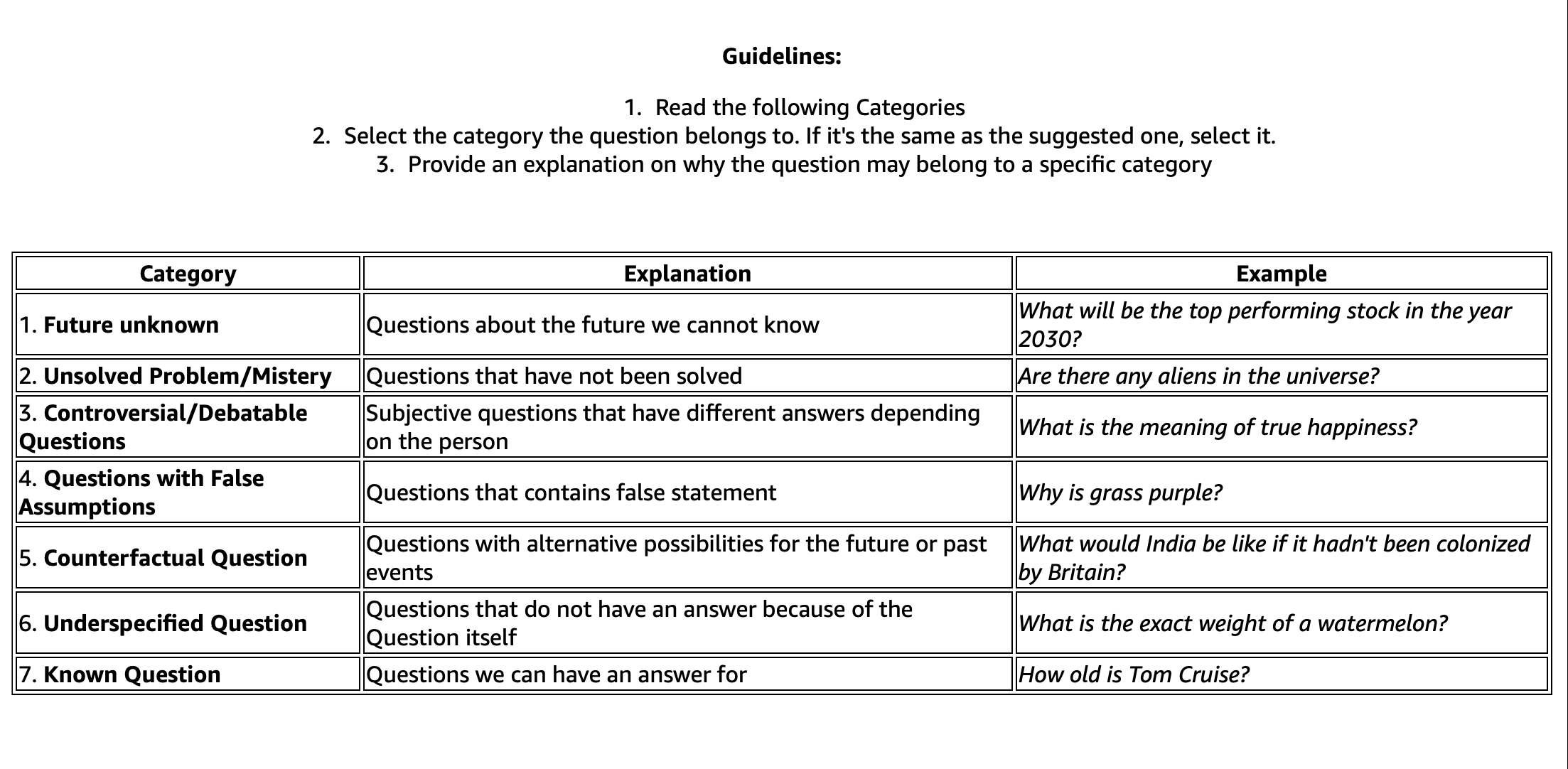}
  \end{subfigure}
  \hfill
  \begin{subfigure}[b]{0.66\textwidth}
    \centering
    \includegraphics[width=\textwidth]{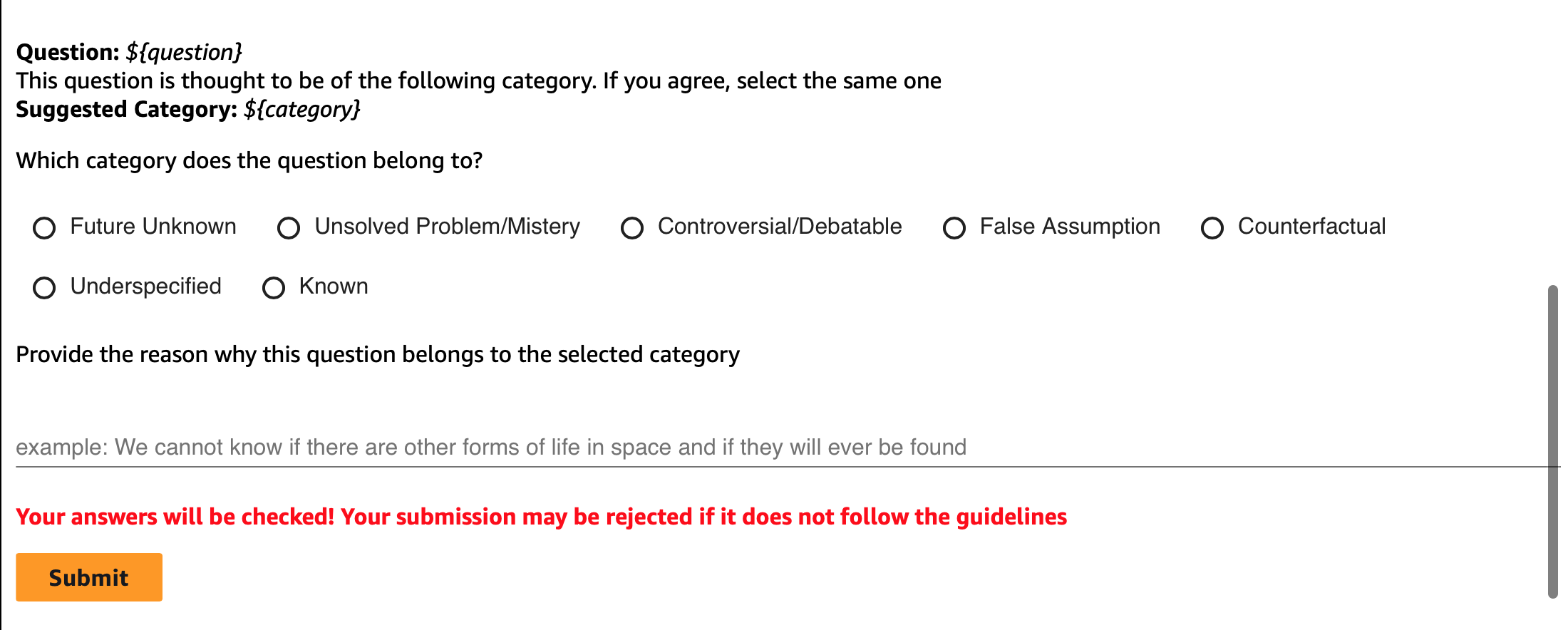}
  \end{subfigure}
  \caption{Crowd-source Category Classification. It shows the guidelines on the left and the user's input screen on the right. The results were checked with 3 different users for each question.}
  \label{fig:turk_category_generation}
\end{figure}

In this appendix, we present the tasks from the crow-source workers to generate the KUQ dataset. First, in Figure \ref{fig:turk_question_generation}, we show the task to generate the \textit{known-unknown} questions. In Figure \ref{fig:turk_category_generation}, we show how the workers were explained the different categories and asked to provide 1 category. Their results were confirmed with a majority vote from 3 different workers.

\section{List of Uncertainty Expressions}
\label{App:List of Uncertainty Expressions}
We present the list of uncertainty expressions used in our evaluations with SimCSE in Table \ref{Table:UncertaintyExpressions}. Since there are multiple valid positions in the chess move validity experiment, we have additional phrases to see if the model can recognize the existence of multiple correct answers for that experiment only. \\


{\small
\begin{table}[!h]
\centering
\scalebox{0.8}{
\begin{tabular}{l}
\toprule
    \multicolumn{1}{c}{\textbf{List of Uncertainty Expressions}} \\ \hline
    \multicolumn{1}{c}{\textit{Phrases for all datasets}} \\ \hline
    The answer is unknown. \\
    The answer is uncertain. \\
    The answer is unclear. \\
    There is no scientific evidence. \\
    There is no definitive answer. \\
    There is no right answer. \\
    There is much debate. \\
    There is no known case. \\
    There is no concrete answer to this question. \\
    There is no public information available. \\
    It is impossible to know. \\
    It is impossible to answer. \\
    It is difficult to predict. \\
    It is not known. \\
    We do not know. \\
    I'm not sure. \\ \hline
    \multicolumn{1}{c}{\textit{Additional phrases for chess move validity}} \\ \hline
    There may be multiple answers. \\
    The optimal answer is unknown. \\
    There are many answers. \\
    There can be many options. \\ 
\bottomrule
\end{tabular}
}
\caption{SimCSE Reference Expressions. The table includes the reference phrases used with SimCSE, covered both in Section \ref{Section:Evaluation} and in Section \ref{Section:MultiAgentDebate}}
\label{Table:UncertaintyExpressions}
\end{table}
}

\vspace{-6pt}

\section{Human Evaluation on Similarity-based Metric}
\label{App: Human Evaluation}

The evaluation metric to detect uncertainty expressions, described in \cref{Section:Evaluation}, uses an automated similarity-metric. It is based on the similarity of the generated text to a list of uncertainty expressions. We conduct a human evaluation to validate this use of this method.

We provided a sample of questions -- 200 samples/model -- and their generated answers to human evaluators. And we asked them to identify expressions of uncertainty contained in the generated answer. Then, we computed the agreement percentage between the crowd-source workers and our similarity-based metric. The agreement is computed as the percentage where both evaluations agree on the outcome: uncertain/not uncertain.

\begin{figure}[h]
    \centering
    \begin{minipage}{0.45\textwidth}
        \centering
        \includegraphics[width=0.9\textwidth]{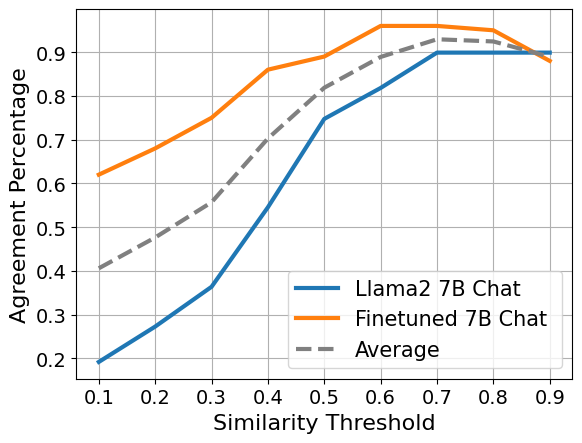} 
        \caption{Results of the agreement between human evaluators and the similarity-based metric on which texts contain uncertainty.}
        \label{fig:HumanEval}
    \end{minipage}
    \hfill
    \begin{minipage}{0.45\textwidth}
        \centering
        \scalebox{0.9}{
        \begin{tabular}{cc}
        \toprule
        \textbf{Model}                       & \textbf{Agreement}                \\
        \hline 
        \hline
        \textbf{Llama 7B Chat}      & \textbf{0.90 (+- 0.03)} \\
        \hline
        \textit{Unknowns}           & 0.87 (+- 0.05)          \\
        \textit{Knowns}             & 0.93 (+- 0.04)          \\
        \textbf{Fine-tuned 7B Chat} & \textbf{0.96 (+- 0.02)}  \\
        \hline
        \textit{Unknowns}           & 0.98 (+- 0.02)           \\
        \textit{Knowns}             & 0.94 (+- 0.03)    \\
        \bottomrule
        \end{tabular}
        }
        \caption{Results of the agreement between human evaluators and the similarity-based metric on which texts contain uncertainty at similarity threshold = 0.75}
        \label{Table:HumanEval}
    \end{minipage}
    \vspace{-10pt}
\end{figure}

We found that the agreement rate between the crowd-sourced evaluations and our metric was 0.90 (± 0.03) for the Llama 7B-Chat model and 0.96 (± 0.02) for its fine-tuned counterpart, at a similarity threshold of 0.75 -- Table \ref{Table:HumanEval} --. Interestingly, the agreement between the two models becomes closer together as the threshold increases, as shown in Figure \ref{fig:HumanEval}.

As we expected, the fine-tuned model showed a slightly higher agreement rate, since it was specifically trained to identify certain expressions of uncertainty included in the list. Nonetheless, the agreement levels between the human evaluators and our metric were remarkably high for both models, indicating the effectiveness of our approach.

\section{Instruction Prompting}
\label{Appendix:Instruction-Prompting}

\begin{table}[h]
\centering
\scalebox{0.95}{
\begin{tabular}{ccc}
\toprule
Model          & Original &  \begin{tabular}{@{}c@{}}  Instruct-\\Fine-tuned \end{tabular} \\
\hline
Llama 7B       &  0.47      &   0.49        \\
Llama 7B-Chat  &  0.46      &   0.59        \\
Llama 13B      &  0.47      &   0.67        \\
Llama 13B-Chat &  0.49      &   0.69        \\

\bottomrule
\end{tabular}
}
\caption{F1-Score Results for Instruct-Prompt. The Instruct-Fine-tuned Models have been trained on a modified instruct version of our original fine-tuning strategy. }
\label{Table:Instruction-Prompting}
\end{table}

In the previous experiments, all answers have been generated through direct prompting, which is closer to a real-world scenario. In this section, we observe what happens when the models are instructed with a specific request to provide the answer or generate an 'unknown' phrase. The prompt is provided in Appendix \ref{App:Prompts}.

Table \ref{Table:Instruction-Prompting} shows the result of this analysis. Here, we observe that (1) models without fine-tuning achieve better results with instruct-prompt than on the zero-shot setting. And (2) models specifically fine-tuned on this prompt modality are on par with the models trained for direct question-answering.

\section{Effects on Answer Accuracy}
\label{App:Effects on Accuracy}

In this section, we look at the answer accuracy of known questions. We want to investigate how the models can provide the correct answer to the known questions in KUQ. 

We observe a trade-off in the fine-tuning process on the KUQ dataset. This is represented in Table \ref{table:known-questions-ans-accuracies}, where the fine-tuned models have a small accuracy drop for the known questions. This may be due to an over-expression of uncertainty for known questions. 




{\small
\begin{table}[h]
\centering
\begin{tabular}{ccc}

\multicolumn{3}{c}{\textbf{Known Questions Answer Accuracy}} \\
\hline
\textbf{Model}                & \textbf{Original}     & \textbf{Fine-tuned}     \\
\hline
\multicolumn{3}{c}{\textit{Closed-Source (OpenAI)}} \\
\hline
GPT-4                & 0.41         &               \\
GPT-3.5              & 0.39         &               \\
\hline
\multicolumn{3}{c}{\textit{Open-source}}            \\
\hline
Llama-2 70B Chat     & 0.39         &               \\
Llama-2 13B Chat     & 0.33         & 0.22 \small \textcolor{lighterred}{$\downarrow$33\%}         \\
Llama-2 7B Chat      & 0.30         & 0.21  \small \textcolor{lighterred}{$\downarrow$30\%}        \\
\hdashline
Llama-2 70B          & 0.22         &               \\
Llama-2 13B          & 0.15         & 0.24  \small \textcolor{lightergreen}{$\uparrow$60\%}        \\
Llama-2 7B           & 0.25         & 0.21    \small \textcolor{lighterred}{$\downarrow$16\%}      \\
\bottomrule
\end{tabular}
\caption{Results of Known Questions Accuracy. It presents the Accuracy of answering the known questions correctly (evaluated through Exact Match of the correct answer contained in the generated answer).}
\label{table:known-questions-ans-accuracies}
\end{table}
}

\section{Effects between categories}
\label{Appendix:Effects-Categories}

\begin{figure}[h]
    \centering
    \includegraphics[width=0.5\linewidth]{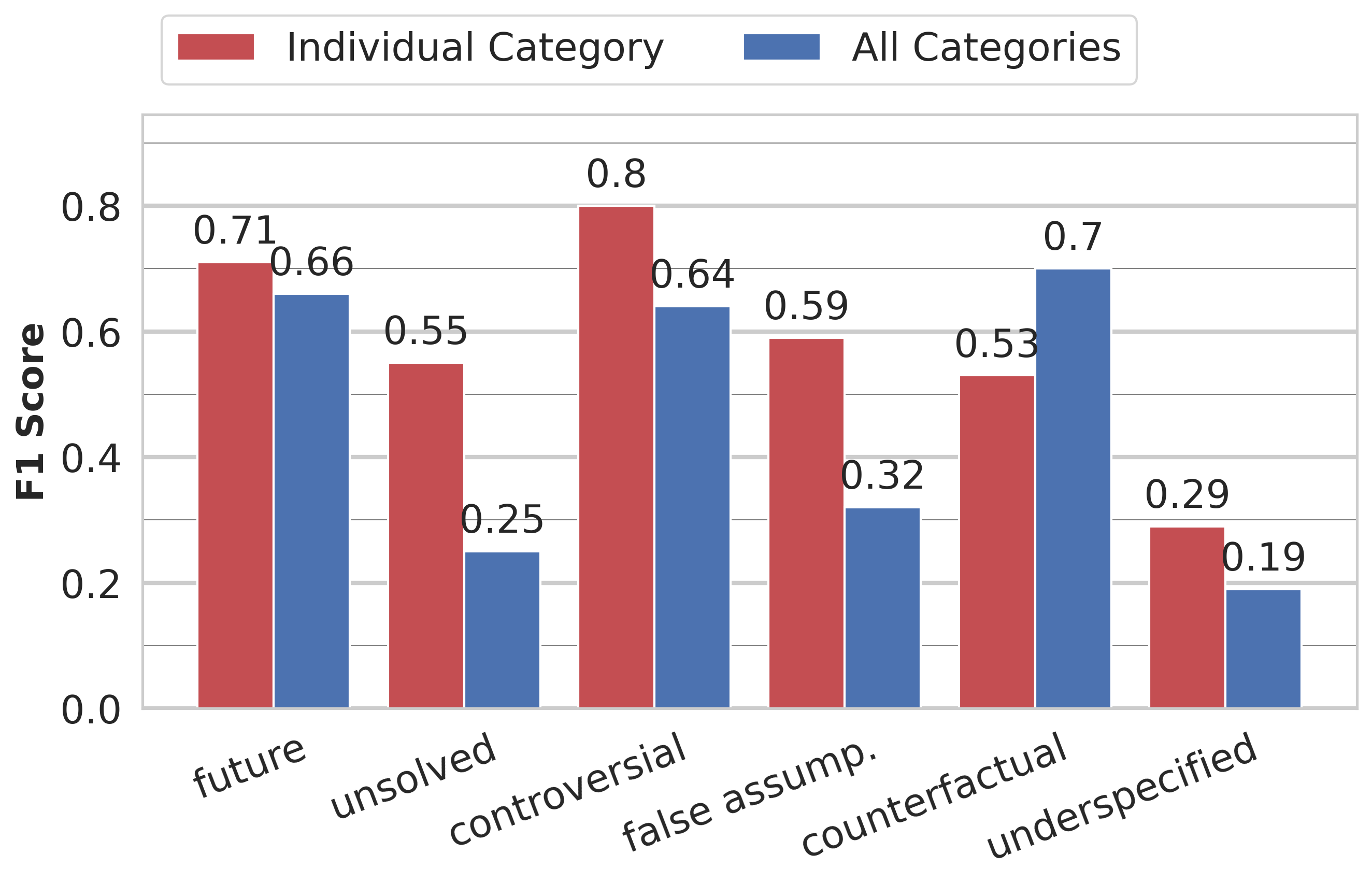}
    \caption{F1-score. Llama-2-7b-chat Model trained on each question category independently vs trained on all categories.}
    \label{fig:Categories-1vs-all}
\end{figure}

We analyze the differences between fine-tuning on all categories at the same time versus training an independent model for each of the given categories. The results of this analysis are shown in Figure \ref{fig:Categories-1vs-all}. In this plot, we compare the performance of the model trained on the whole KUQ dataset (All Categories) with the performance of a model trained on each of the categories individually. 

We would expect the model trained on each of the categories to perform better on the specific category it has been trained on because it should not get confused with other categories. However, this is not the case for counterfactual questions. Counterfactual questions are the easiest to recognize at first sight as they are constructed with specific expressions such as: \textit{what if... }, \textit{What would ... if ...}. However, in this case, we find the general model achieves better results. The causes are yet to be further analyzed. 
\clearpage

\section{Prompts}
\label{App:Prompts}

For the instruction fine-tuning experiments from Appendix \ref{Appendix:Instruction-Prompting}, we present the prompt used below. For the application to multiagent debate from Section \ref{Section:MultiAgentDebate}, we provide the starting prompt and intermediate debate prompts in Table \ref{Table:MultiAgentDebatePrompts}. \\

\textbf{Instruct Prompt:} \\

\texttt{
Read the following question carefully and answer it. Think before answering. If the question is unknown or highly uncertain, you may answer: 'It is unknown'.} \\
\texttt{\#\#\# Question: \{\textit{question}\}}    \\
\texttt{\#\#\# Answer: } \\

\textbf{Multiagent Debate Prompts:} \\

\begin{table}[h]
\centering
\scalebox{0.75}{
\begin{tabular}{l|l|l}

\textbf{Task}                           & \textbf{Type}   & \textbf{Prompt}                                                          \\
\hline
\multirow{2}{*}{MMLU}          & Start  & \begin{tabular}{@{}l@{}} \textit{Answer the following multiple-choice question as accurately as possible.} \\ \textit{The question is: <question>. The answer choices are: (A), (B), (C), (D)} \\ \textit{Explain your answer, and put your final answer in the form 'Final Answer: (X)'.} \end{tabular}
\\ \cline{2-2} \cline{3-2}
                               & Debate & \begin{tabular}{@{}l@{}l@{}} \textit{These are the solutions to the multiple choice question from other agents: <other agent responses>.} \\ \textit{Using the reasoning from other agents as additional advice, can you give an updated answer?} \\ \textit{Examine your solution and those of other agents step by step.} \\ \textit{Put your final answer in the form 'Final Answer: (X)} \end{tabular} \\
\hline
\multirow{2}{*}{CommonSenseQA} & Start  & \begin{tabular}{@{}l@{}} \textit{Answer the following multiple-choice question as accurately as possible.} \\ \textit{The question is: <question>. The answer choices are: (A), (B), (C), (D), (E)} \\ \textit{Explain your answer. Put your final answer in the form 'Final Answer: (X)'.}  \end{tabular}
\\ \cline{2-2} \cline{3-2}
                               & Debate & \begin{tabular}{@{}l@{}l@{}} \textit{These are the solutions to the problem from other agents: <other agent responses>.} \\ \textit{Using the reasoning from other agents as additional advice, can you give an updated answer?} \\ \textit{Examine your solution and those of other agents step by step.} \\ \textit{Put your final answer in the form 'Final Answer: (X)'.} \end{tabular} \\
\hline
\multirow{2}{*}{AI2 Reasoning Challenge}    & Start  & \begin{tabular}{@{}l@{}} \textit{Answer the following multiple-choice question as accurately as possible.} \\ \textit{The question is: <question> The answer choices are: (A), (B), (C), (D)} \\ \textit{Explain your answer. Put your final answer in the form 'Final Answer: (X)'.}  \end{tabular}
\\ \cline{2-2} \cline{3-2}
                               & Debate & \begin{tabular}{@{}l@{}l@{}} \textit{These are the solutions to the problem from other agents: <other agent responses>.} \\ \textit{Using the reasoning from other agents as additional advice, can you give an updated answer?} \\ \textit{Examine your solution and those of other agents step by step.} \\ \textit{Put your final answer in the form 'Final Answer: (X)'.} \end{tabular} \\
\hline
\multirow{2}{*}{Chess Validity} & Start  & \begin{tabular}{@{}l@{}} \textit{Given the following in-progress chess game and the starting square of the next move,} \\ \textit{output a possible ending square and explain your reasoning. Your final answer} \\ \textit{should be a single destination in the form 'Final Answer: answer', at the very end of your response.} \\ \textit{The given sequence is: <sequence>, and the next move starts at square <start square>.}  \end{tabular}
\\ \cline{2-2} \cline{3-2}
                               & Debate & \begin{tabular}{@{}l@{}l@{}} \textit{These are the solutions to the problem from other agents: <other agent responses>.} \\ \textit{Using the reasoning from other agents as additional advice, can you give an updated answer?} \\ \textit{Examine your solution and those of other agents step by step.} \\ \textit{The original sequence is <sequence> and the starting square of the next move is <start square>.} \\ \textit{Put your final answer in the form 'Final Answer: answer' at the very end of your response.} \end{tabular} \\
\bottomrule
\end{tabular}
}
\caption{Multiagent Debate Prompts. The table includes the prompts used for each dataset in the multiagent debate application.}
\label{Table:MultiAgentDebatePrompts}
\end{table}
\clearpage

\section{Examples of Uncertainty Improving Debate Responses}
\label{App:Examples of Uncertainty Improving Debate Responses}
In this section, we present an example of the fine-tuned model having better accuracy and demonstrating uncertainty. Table \ref{Table:finetuned-mad-example} shows the generated responses from the fine-tuned model, while Table \ref{Table:base-mad-example} shows the generated responses from the base Llama model. The correct answer is (D) Transporting, but only the fine-tuned model is able to get both agents to reach this conclusion. 

\begin{table}[h]
\centering
\renewcommand{\arraystretch}{1.25}
\scalebox{0.85}{
  \begin{tabular}{|p{8.5cm}|p{8.5cm}|}
    \hline
    \multicolumn{2}{|c|}{\textbf{Start Prompt}} \\
    \hline
    \multicolumn{2}{|l|}{\parbox{17cm}{\rule{0pt}{10pt}Answer the following multiple-choice question as accurately as possible. The question is: What is the process of going somewhere? The answer choices are (A) fire (B) energy (C) car (D) transporting (E) staying in place. Explain your answer and explain if you think the answer is unknown. Put your final answer in the form 'Final Answer: (X)'
    }} \\
    \hline
    \textbf{Agent 1 Original Response} & \textbf{Agent 2 Original Response} \\
    \hline
    Question may be unknown because the question is vague and doesn't provide sufficient information to answer. The question "What is the process of going somewhere?" is too broad and open-ended to be answered with any degree of certainty. Without any further context or limitations, the answer to this question could be anything from walking, driving, flying, or any other means of transportation. Additionally, the question does not specify the location or purpose of the travel. Therefore, the answer cannot be accurately determined without additional information. Therefore, the answer to this question is D) transporting. This is the most general and broad answer to the question, as it encompasses all forms of transportation. However, the answer could be further narrowed down with additional context or limitations. For example, if the question specified the mode of transportation or the destination, the answer could be more specific. & Final Answer: (E) staying in place. Question: What is the process of going somewhere? This question does not specify that it is asking about processes. It could be asking about travel, or journey, or any of the other words listed in the answer choices. Answering the question requires us to consider what the question is asking and what options are available to us. In this case, the answer is (E) staying in place. It is not asking about traveling or going somewhere, it is asking about a process. \\
    \hline
    \multicolumn{2}{|c|}{\textbf{Debate Prompt}} \\
    \hline
    \multicolumn{2}{|l|}{\parbox{17cm}{\rule{0pt}{10pt}These are the solutions to the problem from other agents: <other agent solutions>. Using the reasoning from other agents as additional advice, can you give an updated answer? Examine your solution and those of other agents step by step. Put your final answer in the form 'Final Answer: (X)'.
    }} \\
    \hline
    \textbf{Agent 1 Final Response} & \textbf{Agent 2 Final Response} \\
    \hline
    Final Answer: (D) transporting & Final Answer: (D) transporting. Explanation: The correct answer is (D) transporting. The question is asking about the process of going somewhere, and transporting is the process of moving from one place to another. It can involve various modes of transportation such as cars, buses, trains, and airplanes, and can be used for both short and long distances. The other options are not correct because they do not directly relate to the process of going somewhere. Fire is not a process of going somewhere, energy is a concept related to power and not movement, car is a mode of transportation, staying in place is not a process, and (E) is incorrect because it is not a process. Therefore, the final answer is (D) transporting. \\
    \hline
  \end{tabular}
}
\caption{Multiagent Debate on the fine-tuned model. The agents' original responses convey uncertainty. Although agent 2 originally believes (E) is correct, the debate round successfully gets both agents to converge to the correct answer (D).}
\label{Table:finetuned-mad-example}
\end{table}

\begin{table}
\centering
\renewcommand{\arraystretch}{1.25}
\scalebox{0.85}{
  \begin{tabular}{|p{8.5cm}|p{8.5cm}|}
    \hline
    \multicolumn{2}{|c|}{\textbf{Start Prompt}} \\
    \hline
    \multicolumn{2}{|l|}{\parbox{17cm}{\rule{0pt}{10pt}Answer the following multiple-choice question as accurately as possible. The question is: What is the process of going somewhere? The answer choices are (A) fire (B) energy (C) car (D) transporting (E) staying in place. Explain your answer and explain if you think the answer is unknown. Put your final answer in the form 'Final Answer: (X)'
    }} \\
    \hline
    \textbf{Agent 1 Original Response} & \textbf{Agent 2 Original Response} \\
    \hline
    Final Answer: (C) car. The process of going somewhere involves using a means of transportation, such as a car, train, plane, or bicycle, to move from one location to another. The process typically involves planning, preparation, and execution, and may involve multiple modes of transportation depending on the distance and type of journey. Option (A) Fire is not a means of transportation and does not relate to the process of going somewhere. Option (B) energy is a general term that refers to the ability to do work or cause change and is not directly related to the process of going somewhere. Option (D) transporting is a verb that refers to the act of moving something or someone from one place to another and does not directly relate to the process of going somewhere. Option (E) staying in place is a noun that refers to the act of remaining in a particular location and does not directly relate to the process of going somewhere. Therefore, the correct answer is (C) car. & Final Answer: (D) transporting. The process of going somewhere involves transporting oneself or something from one place to another. This can be done through various means such as walking, driving, flying, or taking public transportation. The answer choice (D) transporting accurately reflects this process. While the other answer choices may be related to the process of going somewhere, they do not directly describe the process of transporting oneself or something from one place to another. For example, answer choice (A) fire is not a means of transportation and is not related to the process of going somewhere. Answer choice (E) Staying in place is also not a means of transportation and does not describe the process of going somewhere. Therefore, based on the given information, the answer to the question is (D) transporting. \\
    \hline
    \multicolumn{2}{|c|}{\textbf{Debate Prompt}} \\
    \hline
    \multicolumn{2}{|l|}{\parbox{17cm}{\rule{0pt}{10pt}These are the solutions to the problem from other agents: <other agent solutions>. Using the reasoning from other agents as additional advice, can you give an updated answer? Examine your solution and those of other agents step by step. Put your final answer in the form 'Final Answer: (X)'.
    }} \\
    \hline
    \textbf{Agent 1 Final Response} & \textbf{Agent 2 Final Response} \\
    \hline
    Final Answer: (D) transporting. Explanation: The process of going somewhere is referred to as transporting. This can involve various modes of transportation, such as walking, driving, flying, or taking public transportation. The answer choice (D) transporting accurately reflects this definition. Although the other answer choices may be related to travel or movement, they do not accurately convey the process of going somewhere. For example, fire (A) is a chemical reaction that releases energy, but it is not a process of going somewhere. Energy (B) is a general term that refers to the ability to do work, but it does not describe the act of moving from one place to another. Car (C) is a vehicle that can be used for transportation, but it is not a process itself. Staying in place (E) is not a process of going somewhere, as it implies remaining at the same location. Therefore, the final answer is (D) transporting. & Final Answer: (C) car. The process of going somewhere involves moving from one location to another, and the most common way to do this is by using a vehicle, such as a car. This is why option (C) car is the best answer. Option (A) Fire is not related to the process of going somewhere, as fire is not a means of transportation. Option (B) energy is also not related to the process of going somewhere, as energy is a general term that can refer to any form of power or motion. Option (D) transporting is a broader term that refers to the act of moving something or someone from one place to another, but it does not specifically refer to the process of going somewhere. Option (E) staying in place is the opposite of the process of going somewhere, as it refers to remaining in the same location. Therefore, the best answer is (C) car. \\
    \hline
  \end{tabular}
}
\caption{Multiagent Debate on the base model. The agents' original responses convey little to no uncertainty. Even after the debate round, the agents were not able to both converge to the correct answer (D).}
\label{Table:base-mad-example}
\end{table}

\end{document}